\RequirePackage{fix-cm}
\documentclass[twocolumn]{svjour3}
\smartqed

\makeatletter
\let\cl@chapter\undefined
\makeatletter

\usepackage[pagebackref,colorlinks,linkcolor=red,anchorcolor=blue,citecolor=green,CJKbookmarks=True]{hyperref}
\usepackage{amsmath,amssymb,amsfonts}
\usepackage{url}
\usepackage{booktabs}
\usepackage{graphicx}
\usepackage{natbib}
\usepackage{bbding}
\usepackage{multirow}
\usepackage{booktabs}
\usepackage{dsfont}
\usepackage{times}

\newcommand{\ie}{\emph{i.e.}}
\newcommand{\eg}{\emph{e.g.}}
\newcommand{\etc}{\emph{etc.}}

\def\Fix#1{{#1}}
\def\FixCite#1{{#1}}

\begin{document}

\title{Instance-Level Moving Object Segmentation from a Single Image with Events
}

\author{Zhexiong Wan \and
        Bin Fan \and
        Le Hui \and
        Yuchao Dai~\textmd{\Envelope} \and
        Gim Hee Lee
}

\institute{
Zhexiong Wan \and Le Hui \and Yuchao Dai (Corresponding author) \at
School of Electronics and Information, Northwestern Polytechnical University and Shaanxi Key Laboratory of Information Acquisition and Processing, Xi'an, Shaanxi 710129, China. \\ 
Zhexiong Wan is also with the Department of Computer Science, National University of Singapore, Singapore 117417. \\ 
\email{wanzhexiong@mail.nwpu.edu.cn, huile@nwpu.edu.cn, daiyuchao@nwpu.edu.cn} \\
\and
Bin Fan \at
National Key Laboratory of General Artificial Intelligence, School of Intelligence Science and Technology, Peking University, Beijing 100871, China. \\
\email{binfan@pku.edu.cn} \\
\and
Gim Hee Lee \at
Department of Computer Science, National University of Singapore, Singapore 117417.\\
\email{gimhee.lee@nus.edu.sg}
}

\date{Received: date / Accepted: date}

\maketitle

\begin{abstract}
Moving object segmentation plays a crucial role in understanding dynamic scenes involving multiple moving objects, \Fix{while the difficulties lie in taking into account both spatial texture structures and temporal motion cues.}
Existing methods based on video frames encounter difficulties in distinguishing whether pixel displacements of an object are caused by camera motion or object motion due to the complexities of accurate image-based motion modeling. 
\Fix{Recent advances exploit the motion sensitivity of novel event cameras to counter conventional images' inadequate motion modeling capabilities, but instead lead to challenges in segmenting pixel-level object masks due to the lack of dense texture structures in events. }
\Fix{To address these two limitations imposed by unimodal settings}, we propose the first instance-level moving object segmentation framework that integrates complementary texture and motion cues. 
Our model incorporates implicit cross-modal masked attention augmentation, explicit contrastive feature learning, and flow-guided motion enhancement to exploit dense texture information from a single image and rich motion information from events, respectively.
By leveraging the augmented texture and motion features, we separate mask segmentation from motion classification to handle varying numbers of independently moving objects. 
\Fix{Through extensive evaluations on multiple datasets, as well as ablation experiments with different input settings and real-time efficiency analysis of the proposed framework, we believe that our first attempt to incorporate image and event data for practical deployment can provide new insights for future work in event-based motion related works. }
The source code with model training and pre-trained weights is released at \url{https://npucvr.github.io/EvInsMOS}. 

\end{abstract}

\begin{sloppypar}

\section{Introduction}
\label{sec:intro}

Moving object segmentation (MOS) is a long-standing computer vision task that aims to segment moving objects in a dynamic scene.
\Fix{The difficulties lie in how to simultaneously take into account both spatial texture structures and temporal motion cues to determine whether an object is moving or not as well as segmenting pixel-level object mask.}
It has a wide range of applications such as traffic surveillance~\citep{garcia_backgroundsurvey_CSR_2020, cd:patil_multi_PR_2022} and autonomous driving~\citep{huang2019apolloscape}.
Previous works have investigated various input settings that include video frames~\citep{mos:giraldo_graphmos_TPAMI_2020, mos:homeyer_mostrans_ICCVW_2023},
stereo frames~\citep{mos:pan_jointstereomos_TIP_2020, mos:cao_independent_stereoscopic_CVPR_2019},
optical flow~\citep{mos:Meunier_EMMOS_TPAMI_2023}, events~\citep{eventmos:parameshwara_spikems_IROS_2021, eventmos:zhou_EMSGC_TNNLS_2021},
point clouds~\citep{mos:chen_MOSLidar_RAL_2021}, \etc \Fix{In particular, the image-based MOS methods typically input at least two adjacent frames and can further incorporate more images~\citep{cd:patil_multi_PR_2022, mos:Meunier_EMMOS_TPAMI_2023}.
However, these methods are still restricted by insufficient motion modeling capabilities leading to difficulties in distinguishing between different moving objects as well as whether the pixel movements are caused by the camera or the objects. 
}
Consequently, segmenting multiple independently moving objects (IMOs) in complex dynamic scenes remains an open and challenging problem.

\begin{figure}[t!]
\centering
\includegraphics[width=1.0\linewidth]{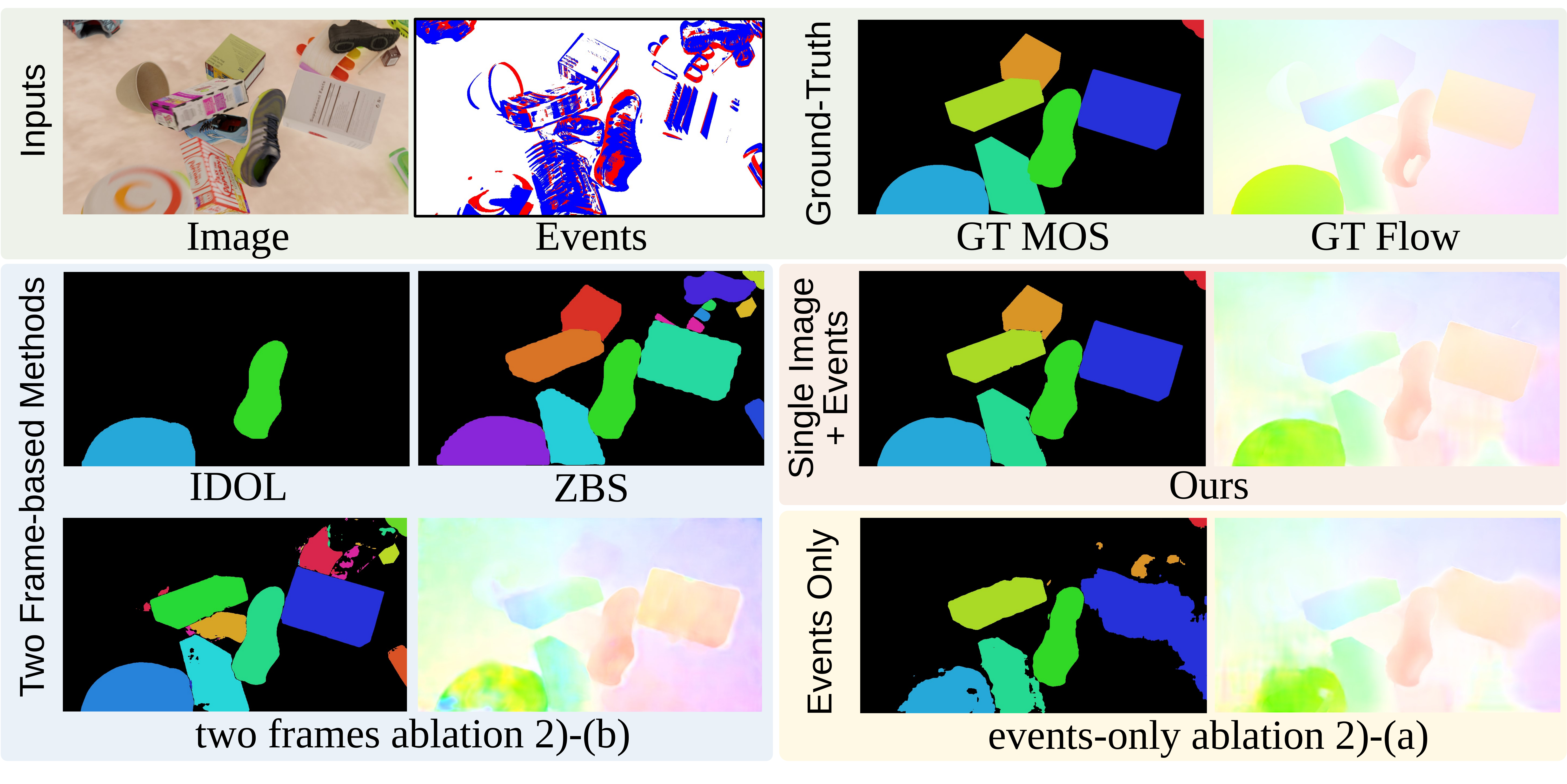} 
\caption{
\textbf{A challenging example of MOS with multiple independently moving objects (IMOs)}. 
\Fix{When multiple static and moving objects coexist in the view of a moving camera, there are two major challenges: 1) static objects projected onto 2D tend to shift relative to the background due to depth parallax, and 2) static objects still present pixel displacements in the image view and also trigger events in the event camera view. 
These make it difficult to distinguish actually moving objects in such complex dynamic scenes.} 
We compare IDOL~\citep{vis:wu_defenseIDOL_ECCV_2022} and ZBS~\citep{cd:an_zbs_CVPR_2023}, which use only image inputs, with our two ablation models that use events only and two images as inputs, respectively. 
The events are visualized with red indicating brightness increase and blue indicating decrease. 
From the comparisons, using only images as input is susceptible to interference from camera motion, which could lead to misjudgment of static objects as moving. 
Additionally, due to the lack of dense texture information in events, moving objects may be misidentified as a single object when they move nearby. 
In contrast, our model can effectively integrate the advantages of the dense texture from image data and the rich motion from event data for accurate segmentation of IMOs. 
} 
\label{viz:teaser}
\end{figure}

In recent years, bio-inspired neuromorphic sensors, such as event cameras, have pioneered a new paradigm for mimicking human vision~\citep{gallego_eventsurvey_TPAMI_2022, zheng_deepeventsurvey_arXiv_2023}. 
Event cameras can capture per-pixel brightness changes asynchronously with high temporal resolution and produce event streams closely related to scene motion, thus opening up new avenues for handling dynamic scenes.
Typically, applications such as optical flow~\citep{eventapp:mostafavi_learnHDR_IJCV_2021, eventflow:shiba_secrets_ECCV_2022} and scene flow estimation~\citep{eventflow:Wan_RPEFlow_ICCV_2023}, object tracking~\citep{eventapp:zhu_evtranstrack_ICCV_2023, eventapp:zhang_universaltrack_IJCV_2023}, motion deblurring~\citep{eventapp:zhang_DMPHN_IJCV_2023, eventapp:zhou_lowdeblurring_IJCV_2023}, depth estimation~\citep{depth:lou_zest_NIPS_2024}, novel view synthesis~\citep{eventapp:low_robustenerf_ICCV_2023}, and video prediction~\citep{eventapp:Zhu_VFPSIE_AAAI_2024} have been investigated, demonstrating the potential of event cameras for applications in dynamic scenes. 
In particular, utilizing event data for MOS has attracted significant interest~\citep{eventdatasets:mitrokhin_EVIMO_IROS_2019, eventdatasets:burner_evimo2_arXiv_2022}. 
Along this vein of sole event inputs, \Fix{traditional motion compensation frameworks}~\citep{eventmos:stoffregen_eventcompensation_ICCV_2019} have been improved by~\cite{eventmos:zhou_EMSGC_TNNLS_2021, eventmos:huang_progressive_CVPR_2023}, and several deep learning-based methods~\citep{eventmos:mitrokhin_learningms_CVPR_2020, eventmos:parameshwara_spikems_IROS_2021} have also been developed. 
\Fix{Recent event-based MOS methods usually use only event data to directly regress the foreground moving mask~\citep{eventmos:parameshwara_spikems_IROS_2021, eventmos:wang_evumoseg_arXiv_2023}. 
However, since events are typically sensitive to areas with brightness changes concentrated in spatially sparse texture structures, the lack of comprehensive dense texture information in events creates difficulties in segmenting a pixel-by-pixel mask with precise contours for each object.
Consequently, the limitation of spatial sparseness still needs to be resolved despite the suitability of applying the event motion-sensitive property to the moving object segmentation task.
}

\Fix{We present a typical example in Fig.~\ref{viz:teaser}, which contains simultaneous movements of multiple objects and the camera motion, to exemplify the above limitations that exist with image-only and event-only unimodal solutions.}
Two frame-based methods~\citep{vis:wu_defenseIDOL_ECCV_2022, cd:an_zbs_CVPR_2023} struggle to cope with multiple moving objects in the presence of camera movement due to the lack of sufficient motion modeling capabilities. 
\Fix{The comparison with the two unimodal ablation models in Tab.~\ref{tab:ablation1}-2) also verifies their respective limitations.
The two-frame based model shown in 2)-(b) has limited motion modeling capabilities and thus fails to exclude some actually static objects when accompanied by camera movements despite being able to segment the spatial contours of the object masks well.
The events-only model shown in 2)-(a) performs better at distinguishing whether an object is moving, but does not perform well in segmenting the contours of the object.
We find that the two modalities have complementary priorities and can leverage the best of both worlds to address these limitations. 
Combining the strengths of both, integrating texture structures in images and motion cues in events meets the needs of the MOS task, which also fits with the modern trend of multimodal data fusion.
}
Specifically, we use \textit{the image which provides rich texture information for dense per-pixel segmentation}, and \textit{the events which encode detailed motion information for precise separation of IMOs} from the static background, static objects, and potential camera movements.

To this end, we present a novel event-enhanced instance-level moving object segmentation (InsMOS) framework that leverages the complementary information between image and events to achieve per-pixel segmentation for all IMOs. 
Specifically, we first augment the extracted image and event features from an implicit perspective by exploiting cross-modal masked attention. 
From an explicit perspective, we impose contrastive learning on texture and motion features based on multi-frame feature consistency \Fix{to ensure these features complement each other}. 
Subsequently, we introduce a flow-guided feature enhancement module to further explicitly reinforce the motion feature learning. 
Unlike some previous methods that segment only the area of foreground moving objects (AMOs) or a fixed number of moving masks, our framework decodes the mask embeddings with corresponding motion scores and selects a variable number of valid segmentation embeddings that satisfy confidence score thresholds to handle multiple instance-level IMOs. 
Through extensive experiments, we validate the effectiveness of combining a single image and events for MOS and demonstrate the state-of-the-art performance achieved by our proposed framework.

Our main contributions are summarized as follows:
\begin{itemize}
\item \Fix{We explore for the first time the complementary characteristics of two modalities, \ie image and event data on the task of moving object segmentation.}
\item \Fix{We propose a novel instance-level moving object segmentation framework that utilizes the texture and motion information from input modalities to learn mask segmentation embeddings and motion confidence embeddings, respectively. This enables instance-level MOS with varying numbers of IMOs. }
\item We introduce an implicit cross-modal masked augmentation module and explicit contrastive feature learning with cross-frame similarity to exploit discriminative complementary texture and motion information. 
\item We develop a tractable flow-guided feature enhancement module to further explicitly refine the motion feature representations derived from input motion observations. 

\end{itemize}

\section{Related Work}
\label{sec:related}

\noindent\textbf{Image-based moving object segmentation} \Fix{aims to segment moving objects from video frames. 
It is a challenging task in computer vision with various applications}
~\citep{mos:tokmakov_learning2seg_IJCV_2019}. 
FgSegNet~\citep{cd:lim_fgsegnet_PRL_2018, cd:lim_fgsegnetv2_PAA_2020} propose to extract multi-scale features for foreground segmentation. 
\cite{mos:pan_jointstereomos_TIP_2020} propose to jointly solve video deblurring, scene flow estimation, and MOS in a unified framework. 
GraphMOS~\citep{mos:giraldo_graphmos_TPAMI_2020} put forward the concepts of graph signal processing for MOS. 
MOVE~\citep{mos:bielski_move_NeurIPS_2022} proposes to segment movable objects from a single image without any form of supervision.
~\cite{objdis:bao_discovering_CVPR_2022} present an auto-encoder-based framework for unsupervised object discovery.
~\cite{objdis:lv_weaklycontrastive_arXiv_2023} incorporate weakly-supervised contrastive learning to enhance texture information.
However, these methods only separate the AMOs from the background, while some recent methods also segment IMOs, \ie, address the InsMOS task.
\cite{mos:cao_independent_stereoscopic_CVPR_2019} propose to segment IMOs by predicting instance-specific 3D scene flow maps and instance masks from stereo videos.
\cite{mos:Meunier_EMMOS_TPAMI_2023} introduce the expectation maximization framework to unsupervised motion segmentation from optical flow. 
ZBS~\citep{cd:an_zbs_CVPR_2023} proposes an instance-level background model based on zero-shot object detection. 
\cite{mos:homeyer_mostrans_ICCVW_2023} propose dual-stream transformers to segment IMOs by fusing image and optical flow. 
DeepFTSG~\citep{cd:rahmon_deepftsg_IJCV_2024} incorporates single and multi-stream network blocks to discriminate salient moving objects. 
Despite the promising results, these methods usually cannot accurately model pixel-level movements, \eg, simultaneous multi-object and camera motion, and thus are not robust to complex multi-object dynamics.

\Fix{\noindent\textbf{Image-based video object segmentation.} 
Image-based video object segmentation (VOS) and video instance segmentation (VIS) have been studied extensively over the years~\citep{zhou_SurveyVideoSeg_TPAMI_2022, gao_deepvosreview_AIR_2023}. 
The tasks of zero-shot VOS (ZVOS, also known as unsupervised VOS) and VIS are also closely related to our InsMOS task. 
The goal of ZVOS is to segment all objects in a video without any guidance (\eg, initial mask or text prompt). 
Early efforts directly rank segments and regress foreground objects~\cite{vos:fragkiadaki2015learning}, with subsequent pixel-level embedding~\citep{vos:li2018instance} and long-term context encoding~\citep{vos:lu2019see} becoming popular. 
}
MATNet~\citep{vos:zhou_motiontrans_AAAI_2020} attempts to incorporate object appearance and motion information. 
Isomer~\citep{vos:yuan_isomer_ICCV_2023} explores the benefit of the long-range dependency modeling capacity of the transformer. 
~\cite{vos:zhao_adaptivezvos_ijcv_2024} presents a multi-source predictor with RGB, optical flow, depth, and saliency map as inputs. 
\Fix{
However, these methods typically only segment salient foreground areas without identifying individual objects, which are not suitable for video analysis that requires fine-grained analysis of varying numbers of objects. 
}
\Fix{Furthermore, the VIS task is more challenging as it requires not only segmenting foreground regions but also separating different object instances. 
RVOS~\citep{vos:ventura2019rvos} takes a recurrent neural network to recover frame-by-frame instances and associate cross-frame instances.
}
VisTR~\citep{vis:wang_endtransformer_CVPR_2021} proposes a direct end-to-end framework built upon transformers. 
~\cite{vis:cheng_mask2vis_arXiv_2021} explore the generalization of the instance segmentation structure Mask2Former~\citep{vis:cheng_mask2former_CVPR_2022} for VIS.
IDOL~\citep{vis:wu_defenseIDOL_ECCV_2022} develops an online framework to reduce the performance gap with the offline paradigm. 
\Fix{Recently, UVIS~\citep{vos:huang_uvis_CVPR_2024} explores the superior generalization and open-set transfer capabilities that visual language models and self-supervised vision models bring to VIS. 
Despite these advances, these methods typically deal with all foreground objects in a scene and lack the direct discriminative ability of whether an object is moving or not. 
In video analysis of complex dynamic scenes including cases where a static object is still accompanied by the same pixel displacements corresponding to the camera, it remains an open challenge to exclude camera motion interference and accurately recognize the attribute of whether an object is moving or not, which is exactly the problem that instance-level moving object segmentation aims to solve. 
}

\noindent\textbf{Event-based motion estimation.} This task is developed to extract motion information from brightness changes in event data~\citep{eventflow:Bardow_simultaneousflowintensity_2016}.
Some early optimization-based methods estimate the motion with events only~\citep{eventmos:Gallego_Unifyingcontrastmax_cvpr_2018, eventapp:gehrig_eklt_IJCV_2020}, but their results mainly focus on the moving boundary rather than the whole motion field because event data are spatially sparse. 
For the same reason, recent learning-based methods~\citep{eventflow:Zhu_EVFlowNet_CVPR_2019,eventflow:Gehrig_DenseRAFTFlow_3DV_2021, eventflow:ponghiran_denceflow_ICCV_2023, eventflow:gehrig_densetimeflow_TPAMI_2024} that use only events to directly regress the optical flow are difficult to produce reliable dense results~\citep{eventflow:Pan_SingleImageFlow_CVPR_2020}. 
Therefore, there is a new trend to combine event data with other modalities, such as images~\citep{eventflow:Lee_Fusion_FlowNet_ICRA_2022, eventflow:Wan_DCEIFlow_TIP_2022}, 
depth~\citep{eventflow:ieng_event3dflow4dsubspace_2017} and point clouds~\citep{eventflow:Wan_RPEFlow_ICCV_2023}, \Fix{to take advantage of} complementary multimodal data. 
Given these successes in introducing events for motion estimation, we explore and fully illustrate the role of combining image and event data for InsMOS.

\begin{figure*}[tbp]
    \centering
    \includegraphics[width=\textwidth]{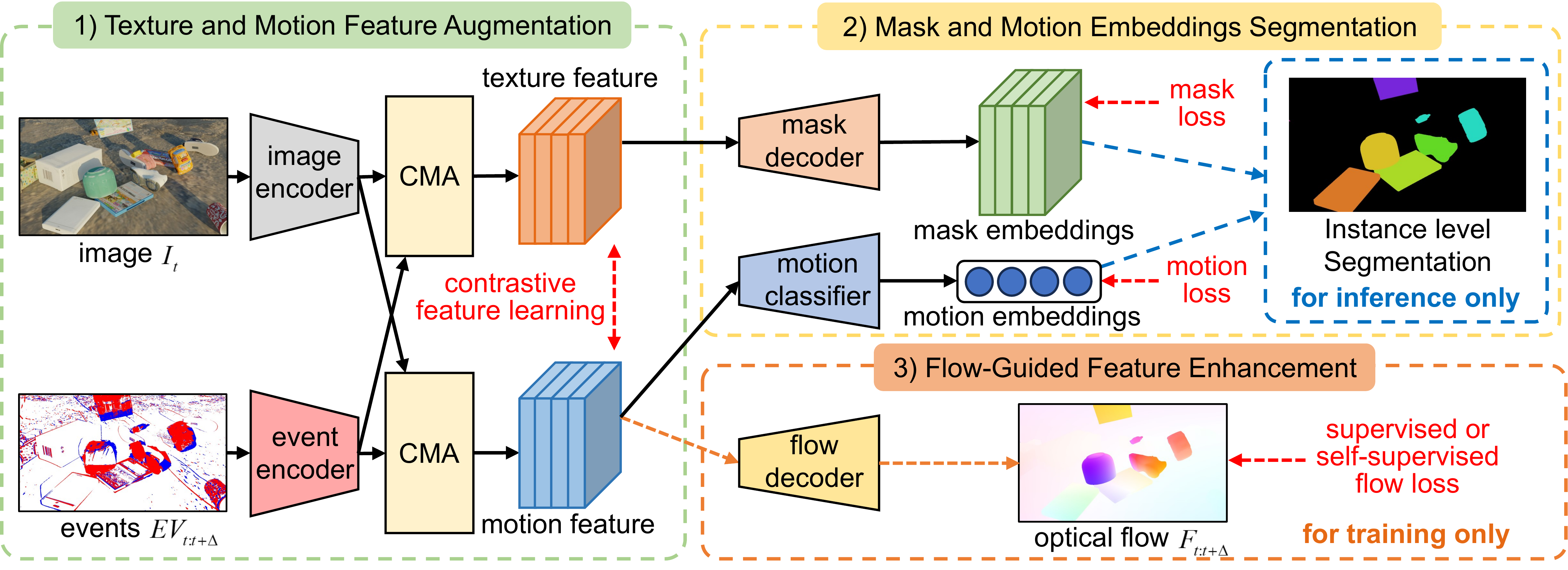}
    \caption{ 
    \textbf{Our proposed InsMOS framework combines a single image with its subsequent events}. 
    The network pipeline is divided into three parts:
    1) The cross-modal masked attention augmentation (CMA) module interactively augments texture and motion representations with an additional contrastive feature learning mechanism applied in training. 
    2) Masks and motion embeddings are decoded separately, allowing for thresholding instance-level segmentation outputs. 
    Since the training loss is applied on full embeddings, this thresholding step only needs to be performed during inference. 
    3) The flow-guided motion feature enhancement module is designed to enhance motion feature learning during training. 
    }  
    \label{fig:framework}
\end{figure*}

\noindent\textbf{Event-based moving object segmentation.} This task aims to segment IMOs from events overlaid with object and camera motion~\citep{eventdatasets:mitrokhin_EVIMO_IROS_2019, eventmos:mitrokhin_modt_iros_2018}. 
The contrast maximization framework~\citep{eventmos:Gallego_Unifyingcontrastmax_cvpr_2018} separates camera motion by maximizing the contrast in warped event images and has been extended to achieve globally optimal motion estimation~\citep{eventmos:peng_globally_TPAMI_2021} and multiple object segmentation~\citep{eventmos:stoffregen_eventcompensation_ICCV_2019}. 
EMSGC~\citep{eventmos:zhou_EMSGC_TNNLS_2021} utilize a spatiotemporal graph representation of events and employ graph cuts for segmentation.
\FixCite{\cite{eventmos:chen_progressivemotionseg_AAAI_2022}} introduce a progressive motion segmentation framework that jointly optimizes motion estimation and event denoising. 
Recently, learning-based methods usually directly regress the AMOs~\citep{eventdatasets:mitrokhin_EVIMO_IROS_2019, eventmos:mitrokhin_learningms_CVPR_2020, eventmos:wang_evumoseg_arXiv_2023}, so they can only determine which positions are moving but cannot separate different IMOs. 
\cite{eventmos:parameshwara_0mms_ICRA_2021} propagate object motion through tracklet decomposition and cluster merging. 
RENet~\citep{eventmos:zhou_MOD_ICRA_2023} fuses images and events to detect moving objects, predicting object bounding boxes rather than pixel-level segmentation. 
JSTR~\citep{eventmos:zhou_jstr_ICRA_2024} proposes to incorporate IMU data and treat events as spatial and temporal 3D point clouds represented by 2D coordinates and timestamps. 
\cite{eventmos:jiang_semantic_VISAPP_2024} propose incorporating semantic and 3D depth information for MOS, while \cite{eventmos:Stamatios_GeneralMOS_3DV_2024} present a multi-stage divide-and-conquer framework based on ego-motion and optical flow estimation.

Previous event-based MOS methods only utilize events as input, potentially leading to the incorrect identification of nearby multiple objects as a single large object due to the absence of texture information in events.
In contrast, we propose for the first time an InsMOS framework to segment all IMOs by exploiting the texture and motion information in a single image and events. 

\section{Method}
\label{sec:method}

Our goal is to utilize both the input image and event observations to extract texture and motion information for precise InsMOS. 
Specifically, we propose two network modules, \ie, cross-modal masked attention augmentation and flow-guided feature enhancement, as well as a contrastive learning mechanism to learn discriminative texture and motion features from both explicit and implicit multi-modal fusion perspectives.
We then separate mask segmentation from motion classification to enable motion segmentation for different object quantities. 
The proposed network architecture is illustrated in Fig.~\ref{fig:framework}, and the details are elaborated below.

\subsection{Texture and Motion Feature Augmentation}

\noindent\textbf{Image and event feature extraction.} 
Unlike the well-known dense tensor structure of conventional images, the event data is an unbounded quaternion ${({e_i})_{i=1}^N} \!=\! {\{ {x_i},{y_i},{t_i},{p_i}\}_{i=1}^N}$, $N$ is the quantity of events. 
Each event consists of the position $(x_i, y_i)$, time $t_i$, and polarity $p_i$ of brightness change.
This special structure leads to the raw events cannot be fed directly into modern CNN-based networks. 
As a result, a common practice is to preprocess raw events into a tensor representation~\citep{eventapp:Rebecq_HighSpeedHDREvent_TPAMI_2019, eventflow:Zhu_EVFlowNet_CVPR_2019}. 
Accordingly, we process a given event slice with an unfixed shape of $N \!\times\! 4$ into a polarity-specific voxel grid~\citep{eventflow:Wan_DCEIFlow_TIP_2022} with a fixed shape of $H \!\times\! W \!\times\! B$: 
\begin{equation}
    EV(x_i, y_i, b) \!=\! \sum_{i=1}^{N}{ p_i \!\max\!{ \left( 0,\! 1 \!-\! \left |b - \!\frac{t_i \!-\! t_{1}}{t_{N} \!-\!  t_{1}}\! (B \!-\! 1) \right| \right) }},
\end{equation}
where $H \!\times\! W$ is the resolution, $B$ is the divided temporal bins, $t_1$ and $t_N$ denote the start and end event timestamps.

For the input image $I_t$ at start time $t\!=\!t_1$ and event voxel $EV_{t:t\!+\!\Delta}$, where $\Delta\!=\!t_N - t$ is the duration of events, we extract the image feature $\mathbf{f}_I$ and event feature $\mathbf{f}_E$ using two backbones without sharing weights. 
Additionally, we introduce event mask encoding as an additional guidance for the model, where $\mathbf{f}_{E\!M}$ is extracted using lightweight three-layer convolutions from the 0-1 binary map with valid events, \ie $EM(x, y) = 1 
~\mathrm{when}~ (x, y) \in (x_i, y_i)^N 
~\mathrm{or}~ 0 ~\mathrm{otherwise}$.

\begin{figure}[t]
    \centering
    \includegraphics[width=\linewidth]{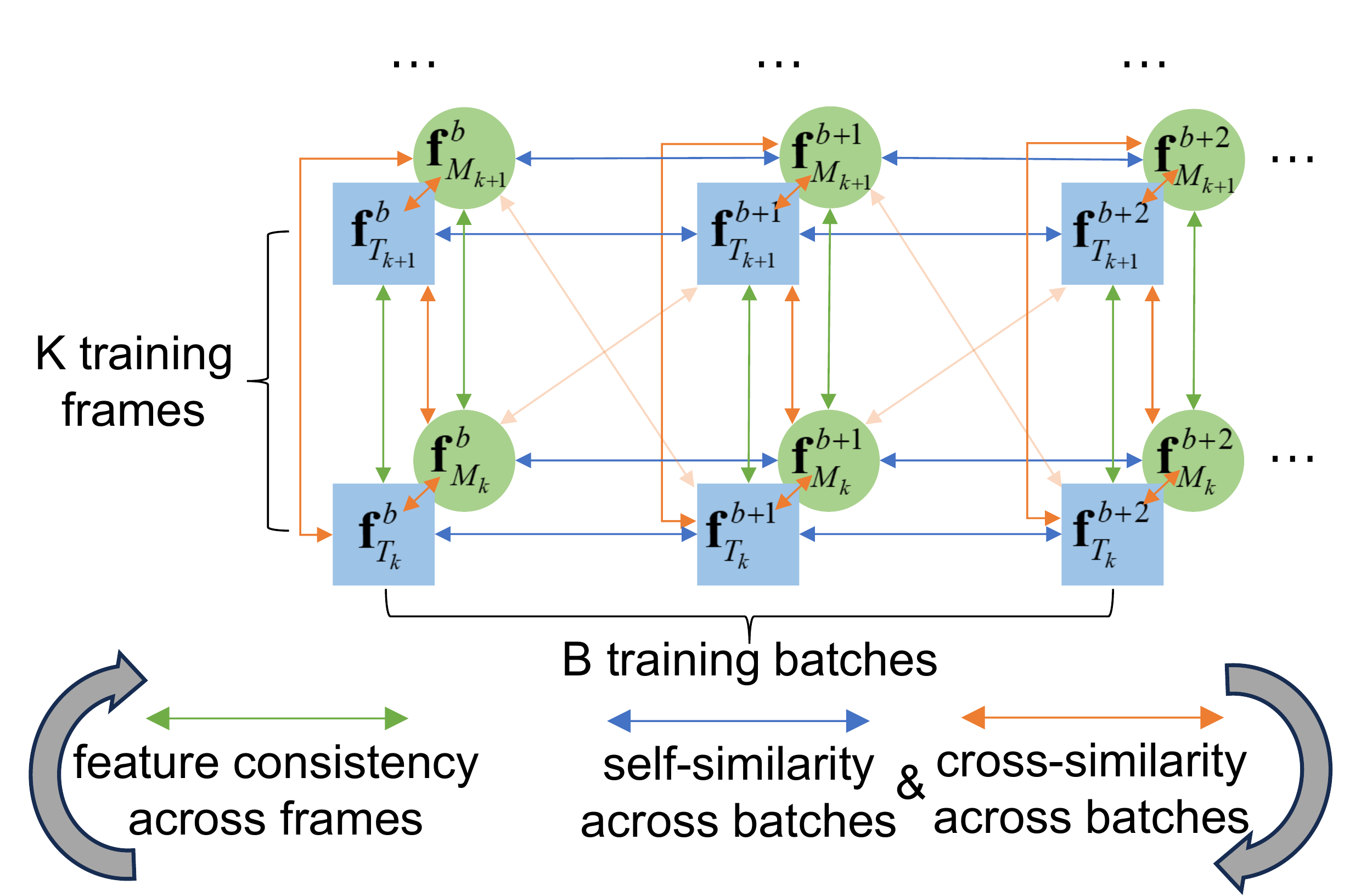}
    \caption{ \textbf{Diagram of our multi-frame contrastive feature learning.} 
    Our feature learning consists of two parts. 
    One is to maximize the feature consistency across frames (green arrows) for the same batch and modality to keep the consistency of the identical modality features. 
    One is to minimize the self-similarity and cross-similarity across batches (blue and orange arrows) to learn complementary information across modalities. 
    The \Fix{bottom-left upward curved arrow} indicates maximizing the consistency measurement to \textbf{reduce the complementarity information} between identical features. 
    While the \Fix{bottom-right downward curved arrow} indicates minimizing the similarity measurements to \textbf{increase the complementary information}.
    Note that only similarities between adjacent batches and frames are drawn for brevity, whereas actually every batch and every frame is considered.
    }  
    \label{fig:closs}
\end{figure}

\vspace{2mm}
\noindent\textbf{Cross-modal masked attention augmentation.}
The texture information embedded in the image feature $\mathbf{f}_I$ is crucial for identifying which pixels belong to objects, while the motion information embedded in the event feature $\mathbf{f}_E$ is important for determining whether pixel displacement is caused by object motion or camera motion. 
Building upon this fundamental insight, we construct a dual-branch cross-modal masked attention augmentation (CMA) module to learn complementary information \Fix{within spatially aligned image and event features}, facilitating the fusion of texture-specific and motion-specific features through two attention branches. 
We adopt the efficient transposed attention (MDTA) module~\citep{zamir_restormer_CVPR_2022} but modify the input from a single image feature to two modalities. 
Additionally, we introduce the event mask feature $\mathbf{f}_{E\!M}$ as an additional masked guidance, multiplying it pixel-by-pixel with the event feature. 
Consequently, the inputs are $\!\mathbf{Q} \!=\! \mathbf{f}_I\!, \mathbf{K} \!=\! \mathbf{V} \!=\! \mathbf{f}_E \mathbf{f}_{E\!M}\!$ in the texture branch and 
$\mathbf{Q} \!=\! \mathbf{f}_E \mathbf{f}_{E\!M}, \mathbf{K} \!=\! \mathbf{V} \!=\! \mathbf{f}_I$ in the motion branch (projections are omitted). 
The cross-feature augmentation process for each branch can thus be expressed as follows:
\begin{equation}
\begin{aligned}
\mathbf{f}_T &= (\mathbf{f}_E \mathbf{f}_{E\!M}) \cdot \mathrm{softmax}{(\mathbf{f}_I^\top \cdot (\mathbf{f}_E \mathbf{f}_{E\!M}) / \tau)} + \mathbf{f}_I, \\
\mathbf{f}_M &= \mathbf{f}_I \cdot \mathrm{softmax}{((\mathbf{f}_E \mathbf{f}_{E\!M})^\top \cdot \mathbf{f}_I / \tau)} + \mathbf{f}_{E}, \\
\end{aligned}
\end{equation}
where $\mathbf{f}_T$ is the augmented texture feature and $\mathbf{f}_M$ is the augmented motion feature, $\cdot$ indicates %
matrix multiplication, 
$\tau$ denotes the learnable temperature coefficient.

\vspace{2mm}
\noindent\textbf{Multi-frame contrastive feature learning.}
To enhance the discriminative capacity of texture feature $\mathbf{f}_T$ and motion feature $\mathbf{f}_M$ through the above augmentation process, we advocate for explicitly modeling the similarity between these two features and reducing their complementarity through contrastive learning. 
To establish the contrastive objective, we propose a multi-frame learning mechanism aimed at intensifying feature similarity across frames while minimizing the similarity between texture and motion features. 
Note that this mechanism operates exclusively during training, the model still only needs a single image and following events as input during inference. 

\Fix{According to InfoNCE~\citep{learning:oord2018representation}}, we first define the similarity measurement $s$ between two features $\mathbf{a}$ and $\mathbf{b}$:
\begin{equation}
\begin{aligned}
s(\mathbf{a},\mathbf{b}) &=\exp (\mathrm{cos}(\mathbf{a},\mathbf{b})/\alpha), \\
\end{aligned}
\end{equation}
where $\mathrm{cos}$ represents the cosine similaritym, and $\alpha$ denotes the learnable temperature coefficient. 

\Fix{
Subsequently, we first compute the feature consistency of identical features across frames $FC_{T}$ and $FC_{M}$ as the maximization objective to fit the inter-frame motion smoothing assumption: 
\begin{equation}
\begin{aligned}
\label{eq:selfandcross1}
FC_{T}^b &= {\textstyle \sum_{k_1, k_2; k_1 \!\ne\! k_2}^{K}} ~ s ({\mathbf{f}^b_{T_{k_1}}}, {\mathbf{f}^b_{T_{k_2}}}), \\
FC_{M}^b &= {\textstyle \sum_{k_1, k_2; k_1 \!\ne\! k_2}^{K}} ~ s ({\mathbf{f}^b_{M_{k_1}}}, {\mathbf{f}^b_{M_{k_2}}}), \\
\end{aligned}
\end{equation}
where $\mathbf{f}_{T_k}^b$ and $\mathbf{f}_{M_k}^b$ are the $k$-th frame texture and motion features at $b$-th batch, $K$ is the number of frames in training. 
$FC_{T}^b$ and $FC_{M}^b$ are the texture and motion feature similarities across all frames at $b$-th batch, respectively. 
We then compute the feature similarity across batches that encompasses self-similarities $SS_{T}, SS_{M}$ and cross-similarity $CS_{T,M}$ across $B$ batches: 
\begin{equation}
\begin{aligned}
\label{eq:selfandcross2}
SS_{T}^b &= {\textstyle \sum_{k}^{K}} ~ {\textstyle \sum_{b'; b' \ne b}^{B}} ~ s(\mathbf{f}_{T_k}^{b}, \mathbf{f}_{T_k}^{b'}), \\
SS_{M}^b &= {\textstyle \sum_{k}^{K}} ~ {\textstyle \sum_{b'; b' \ne b}^{B}} ~ s(\mathbf{f}_{M_k}^{b}, \mathbf{f}_{M_k}^{b'}), \\
CS_{T, M}^b &= {\textstyle \sum_{k_1, k_2}^{K}} ~ {\textstyle \sum_{b'}^{B}} ~ s(\mathbf{f}_{T_{k_1}}^{b}, \mathbf{f}_{M_{k_2}}^{b'}), \\
\end{aligned}
\end{equation}
where $B$ is the training batch size. 
$SS_{T}^b$ and $SS_{M}$ are the texture and motion self-similarities across all other batches at $b$-th batch for all frames, respectively. 
$CS_{T,M}^b$ is the cross-similarity of texture and motion features across all other batches at $b$-th batch for all frames. 
}

In Fig.~\ref{fig:closs}, our feature learning consists of two parts. 
The former is the similarity among features from different batches and modalities across frames, which we take as the denominator to maximize. 
The latter is the cross-frame consistency of features within the same modality and batch, which we take as the numerator of the contrastive objective to minimize. 
\Fix{
The feature learning objective is expressed as: 
\begin{equation}
\begin{aligned}
\label{eq:clloss}
L_{cl} &= - \frac{1}{B} \sum_{b=1}^{B} \log \frac{FC_{T}^b + FC_{M}^b}{SS_{T}^b + SS_{M}^b + CS_{T, M}^b }, \\
\end{aligned}
\end{equation}
where the numerator is the similarity of identical features across different frames as positive pairs in reference to Eq.~\eqref{eq:selfandcross1}.
As in Eq.~\eqref{eq:selfandcross2}, the denominator contains self-similarities of identical features and the cross-similarity of different features across different batches for all frames as negative sample pairs. 
Our principal purpose of minimizing $L_{cl}$ is to maximize the similarity and reduce the between texture features and motion features (\ie~the denominator in Eq.~\eqref{eq:clloss}) to achieve the goal of reducing complementary information and improving descriptiveness. 
}

Notably, this mechanism relies on the assumption of motion smoothing. 
Although actually the texture and motion information constantly evolves over time, and either large or small movements cannot strictly satisfy the smoothing assumption. 
However, the training objective of maximizing the denominator remains unchanged because the difference of the numerator always exceeds the denominator, so the feature learning objective is always set as a regularization term in addition to the data term. 

\subsection{Mask and Motion Embeddings Segmentation}
\label{sec:decoder}
To accommodate the segmentation of varying numbers of IMOs, we divide the segmentation process into two steps: mask segmentation and motion classification. 
The former utilizes the augmented texture feature $\mathbf{f}_T$ to locate pixels corresponding to objects expressed as mask embeddings.
The latter exploits the augmented motion feature $\mathbf{f}_M$ to determine whether an object is moving or stationary,
and outputs the motion embeddings and confidence scores.

\vspace{2mm}
\noindent\textbf{Mask decoder and motion classifier.}
For the motion classification part which solely determines whether each embedding is moving or stationary, \Fix{we utilize a lightweight Feature Pyramid Network (FPN)\FixCite{~\citep{learning:lin2017FPN}} decoder} to produce the motion embeddings $\mathbf{me}_{mov} \!\in \!\mathbb{R}^{c \times n}$ and the motion confidence scores $\mathbf{ms}_{mov} \!\in\! \mathbb{R}^{2 \times n}$ (not depicted in Fig.~\ref{fig:framework} for brevity). 
$n$ is the number of embeddings, and $2$ denotes whether the object is moving (1) or stationary (0).
The fused MOS outputs for all embeddings are obtained through matrix multiplication: 
\begin{equation}
\begin{aligned}
\mathbf{S}_{all}^{pred} \!= \mathbf{me}_{mov}^T \!\cdot \mathbf{me}_{mask} \!\in\! \mathbb{R}^{n \!\times\! h \!\times\! w} \!\Leftarrow\! {\mathbb{R}^{n \!\times\! c}}^T \mathbb{R}^{c \!\times\! h \!\times\! w}.
\end{aligned}
\end{equation}

The training objective involves supervising the fused MOS outputs $\mathbf{S}_{all}^{pred}$ and motion scores $\mathbf{ms}_{mov}$ generated by the mask decoder and motion classifier. 
Initially, we preprocess the ground-truth (GT) segmentation masks of all IMOs in the dataset to obtain the supervision with the same size as the model outputs. 
Since the fixed embeddings number $n$ is larger than the number of moving objects (\ie, GT MOS masks) $m$, we utilize the Hungarian assignment for matching as in semantic segmentation~\citep{cheng_maskformer_NeurIPS_2021}. 
Additionally, we downsample the GT masks to the same size as the decoder output for consistent supervision. 
The processed InsMOS supervision contains the assigned segmentation masks $\mathbf{S}^{gt} \in \mathbb{R}^{n \times h \times w}$ and motion flags $\mathbf{c}^{gt} \in \{0, 1\}^n$, where $\mathbf{c}^{gt}(i)$ is the motion flag (moving 1 or stationary 0) of the $i$-th assigned segmentation mask $\mathbf{S}^{gt}(i)$. 
The MOS task loss is thus formulated as follows: 
\begin{equation}
\begin{aligned}
L_{mos} =& \sum_{i=1}^{n}{\Big[L_{ce}\left(\mathbf{ms}_{mov}(\rho(i)), \mathbf{c}^{gt}(i)\right) + }\\
&\mathds{1}_{\mathbf{c}^{gt}(i)=1} L_{mask}\left( \mathbf{S}^{pred}_{all}(\rho(i)), \mathbf{S}^{gt}(i) \right)\Big],
\end{aligned}
\end{equation}
where $\rho$ denotes the assigned mapping from the Hungarian algorithm. 
$L_{ce}$ represents the binary cross-entropy motion classification loss, \ie, the motion loss in Fig.~\ref{fig:framework}. 
$L_{mask}$ denotes the binary mask loss in Fig.~\ref{fig:framework}, and we employ a weighted combination of focal loss~\citep{lin2017focal} with dice loss~\citep{milletari2016dice} similar in DETR~\citep{carion_DETR_ECCV_2020} to calculate $L_{mask}$. 

\vspace{2mm}
\noindent\textbf{Instance-Level MOS output fusion.}
\label{sec:mosfusion}
The number of fused segmentation predictions $\mathbf{S}_{all}^{pred}$ is a fixed embedding number $n$.
To adapt to varying numbers of segmentation results for $m$ IMOs, we first spatially upsample the predicted masks $\mathbf{S}_{all}^{pred}$ to the full image size $\mathbf{S}_{full}^{pred} \!\in \!\mathbb{R}^{n \times H \times W}$. 
Subsequently, we apply a moving confidence score threshold $\theta$ according to their corresponding motion score $\mathbf{ms}_{mov}$, resulting in:
\begin{equation}
\begin{aligned}
\!\mathbf{S}^{pred} \!&=\! \{\mathbf{S}_{full}^{pred}(i)\}_{i \in [0, n)} ~ \mathrm{if} ~~ \mathrm{softmax}(\mathbf{ms}_{mov}\!)(i) \!\! > \! \!\theta, \!
\end{aligned}
\end{equation}
where the final output $\mathbf{S}^{pred}$ with shape $m \!\times\! H \!\times\! W$ represents the segmentation masks for $m$ IMOs. 
Note that we do not need to perform this process with varying numbers of objects during training (\ie, the blue dashed box in Fig.~\ref{fig:framework}) 
since the task loss is pre-computed on all embeddings $\mathbf{S}_{all}^{pred}$ instead of thresholded embeddings $\mathbf{S}^{pred}$. 

\subsection{Flow-Guided Feature Enhancement}
\label{sec:flowguide}
Our proposed dual-branch augmentation structure with the contrastive feature learning mechanism can model the texture and motion information \textit{explicitly} from events that contain rich motion information.
In addition, we also introduce a flow-guided motion feature enhancement (FFE) module to mine further the motion information in the raw event data \textit{implicitly} since optical flow can directly represent the pixel-by-pixel motion.
Since this module is only used to enhance the motion feature learning in training, we also adopt a lightweight flow decoder with an FPN structure. 
It takes the motion feature $\mathbf{f}_M$ as input and directly predicts the optical flow $F^{pred}_{t:t+\Delta}$. 
To train this module, we perform supervised learning on datasets with GT optical flow labels, or utilize a self-supervised strategy via backward warping for datasets lacking ground-truths. 

For the former, the supervised flow loss is:
\begin{equation}
\begin{aligned}
\label{eq:flowsf}
L_{sf} = \sum_{\mathbf{x}} (\Vert F^{pred}_{t:t+\Delta}(\mathbf{x}) - F_{t:t+\Delta}^{gt}(\mathbf{x})\Vert_2),
\end{aligned}
\end{equation}
where $\mathbf{x}$ are the positions with valid flow ground-truths. 

For the later self-supervised learning, we need to use the two adjacent frames $I_{t}, I_{t+\Delta}$ as a proxy, \ie, 
\begin{equation}
\begin{aligned}
\label{eq:flowuf}
L_{uf} = \sum_{\mathbf{x}} \psi \left[ I_{t}(\mathbf{x}) - I_{t+\Delta}\left(\mathbf{x} + F_{t:t+\Delta}(\mathbf{x})\right) \right], 
\end{aligned}
\end{equation}
where $\psi(x)=(|x|+\epsilon)^q$ is the robust function~\citep{flow:liu_ddflow_aaai_2019} with $\epsilon=0.01$ and $q=0.4$.
Note that the next frame $I_{t+\Delta}$ is only needed during self-supervised training. 

Furthermore, the whole FFE module is disabled when evaluating the segmentation performance (\ie, the orange dashed box in Fig.~\ref{fig:framework}), but it is turned on when visualizing the predicted optical flow or conducting ablation experiments. 

\subsection{\Fix{Training Objective}}

To train our full InsMOS model, we need to balance the weights of the losses from each of the above modules. 
The total loss $L$ is computed as a linear combination of the task loss $L_{mos}$ with the contrastive feature learning objective $L_{cl}$ and flow loss $L_f=L_{sf} ~\mathrm{or}~ L_{uf}$:
\begin{equation}
\begin{aligned}
L = L_{mos} + \lambda_{1} L_{cl} + \lambda_{2} L_f,
\end{aligned}
\end{equation}
where $\lambda_{1}$ and $\lambda_{2}$ are loss combination hyper-parameters.

\begin{table*}[tbp]
\caption{\textbf{Performance comparisons on the eval set of the EVIMO~\citep{eventdatasets:mitrokhin_EVIMO_IROS_2019} dataset}. 
We divided the evaluation into two comparisons: the first focusing on subsequences (top two), and the second on full sequences (bottom two), to ensure fairness. 
SpikeMS can only segment AMOs, thus only the foreground $\rm {mIoU}_{01}$ is calculated. 
EMSGC is a traditional optimization method and we do not count its parameters. 
}
\centering

\setlength{\tabcolsep}{8pt}{
\begin{tabular}{lllllllll}
    \toprule
    Data & Input & Method & $\rm {mAP}$ \Fix{$\uparrow$} & $\rm {mIoU}_{ins}$ \Fix{$\uparrow$} & $\rm {mIoU}_{01}$ \Fix{$\uparrow$} & Time & Param \\
    \midrule
    \multirow{4}*{\rotatebox{90}{subset}} & \multirow{2}*{$E_{t:t\!+\!\Delta} $}
    & SpikeMS & - & - & 16.05 & 49.6ms & \textbf{0.047}M \\
    & & EMSGC & 36.5 & 76.81 & 74.28 & 8.17s & - \\
    \cmidrule{2-8}
    & \multirow{2}*{$\!I_t, \!E_{t:t\!+\!\Delta}$}
    & Ours & 49.7 & 81.55 & 80.52 & 24.8ms & 41.96M \\
    & & Ours-small & 45.6 & 80.29 & 79.30 & 21.0ms & 8.72M \\
    \midrule
    \multirow{4}*{\rotatebox{90}{full}}
    & \multirow{2}*{$\!I_{t}, \!I_{t\!+\!\Delta}\!$}
    & IDOL & 33.2 & 69.72 & 72.02 & 73.1ms & 21.34M \\ 
    & & ZBS & 31.4 & 67.41 & 70.19 & 81.5ms & 175.4M \\ 
    \cmidrule{2-8}
    & \multirow{2}*{$\!I_t, \!E_{t:t\!+\!\Delta}$}
    & Ours & \textbf{36.8} & \textbf{71.99} & \textbf{76.03} & 24.8ms & 41.96M \\
    & & Ours-small & 33.6 & 71.65 & 74.38 & \textbf{21.0ms} & 8.72M \\
    \bottomrule
\end{tabular}
}
\label{tab:evimo}
\end{table*}

\section{Experiments}
\label{sec:experiments}

\subsection{Implementation Details}
\noindent\textbf{Datasets.}
We chose the following two publicly available datasets for our experiments. 
EVIMO~\citep{eventdatasets:mitrokhin_EVIMO_IROS_2019}\footnote{\url{https://better-flow.github.io/evimo/}} is a commonly used dataset specifically designed for event-based motion segmentation. 
It has an official delineation of 53,539 train and 7,862 eval data pairs, each with image and event data captured by a DAVIS camera and corresponding InsMOS labels captured by the motion capture system. 
However, EVIMO suffers from low resolution (only 346$\times$260) and is limited to scenes within laboratory environments, featuring up to 3 IMOs. 
While it provides a large amount of samples for learning-based methods, these limitations hinder thorough validation of the model's performance. 
Consequently, we migrate a recent EKubric~\citep{eventflow:Wan_RPEFlow_ICCV_2023}\footnote{\url{https://npucvr.github.io/RPEFlow/}} dataset simulated for motion estimation. 
This dataset offers motion states of all the objects along with pixel-by-pixel segmentation maps with 19,640 training and 3,824 test pairs.
We generate InsMOS annotations by filtering out non-occluded moving objects based on the position and velocity of each instance. 
Although it contains fewer data samples compared to EVIMO, it has a higher resolution (1280$\times$720) and image quality, featuring more complex and diverse motions (up to 15 IMOs). 
We thus conduct ablation experiments on EKubric to comprehensively illustrate the effectiveness of each component in our proposed model. 
In Fig.~\ref{viz:teaser}, thanks to the detailed metadata from Kubric~\citep{greff2022kubric}, we can exclude the interference of camera motion and accurately determine moving objects. 

\vspace{2mm}
\noindent\textbf{Training details.}
We implement our model with PyTorch on eight RTX3090 GPUs for training, and a single RTX3090 GPU for evaluation. 
We conduct 200k training iterations with a batch size of 16, and Adam optimizer with weight decay $10^{-6}$. 
We employ a one-cycle decay mechanism with the maximum learning rate set to $10^{-4}$. 
All experimental evaluations including comparisons with previous methods are performed on the same evaluation machine. 
The number of event bins is $B\!=\!10$, which is consistent with previous works~\citep{eventapp:Rebecq_HighSpeedHDREvent_TPAMI_2019, eventflow:Wan_DCEIFlow_TIP_2022}. 
The training frames are set to $K\!=\!2$. 
The loss weights are $\lambda_1 \!=\! 1.0$ and $\lambda_2\!=\!2.0$ for numerical scale alignment. 
The moving confidence score threshold is set to $\theta=0.3$, which is consistent with MaskFormer~\citep{cheng_maskformer_NeurIPS_2021}. 
Depending on the maximum number of IMOs, we set the number of embeddings $n$ to 100 for the EKubric dataset and 10 for EVIMO. 

\vspace{2mm}
\noindent\textbf{Metrics.} 
Previous studies~\citep{eventmos:parameshwara_spikems_IROS_2021, eventmos:zhou_EMSGC_TNNLS_2021} typically use the mean Intersection over Union (mIoU) metric to assess the accuracy of segmentation results. 
We calculate the IoU for each moving object and its corresponding prediction, averaging them to obtain $\rm {mIoU}{ins}$, \Fix{where $\rm ins$ denotes that it is an \textbf{ins}tance level MOS eval metric.}
Since some MOS methods can only segment AMOs, we also compute IoU for the moving 0-1 foreground, denoted as $\rm {mIoU}_{01}$. 
However, the IoU metric does not impose penalization of false positives. 
Therefore, we adopt the primary metric of the COCO instance segmentation challenge~\citep{lin2014microsoftcoco}: mean Average Precision $\rm {mAP}$ for a comprehensive InsMOS evaluation. 
Moreover, in our ablation experiments, we introduce optical flow metrics, including end-point-error $\rm {EPE}$ and the ratio of errors less than 1 pixel $\rm {1px}$, to evaluate the motion information modeling capability of these different variants. 

\begin{figure*}[tbp]
\centering
\includegraphics[width=\textwidth]{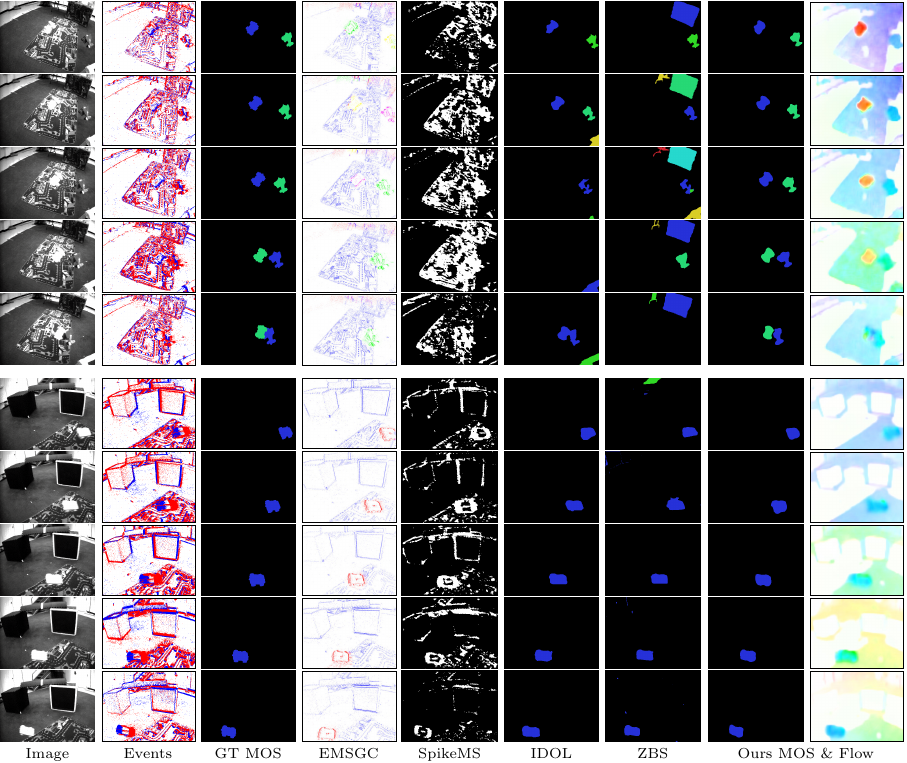}
\caption{\textbf{Visual comparisons on the real EVIMO~\citep{eventdatasets:mitrokhin_EVIMO_IROS_2019} dataset}. 
Our method can segment all IMOs more accurately, especially when they are close together. 
} 
\label{viz:evimo}
\end{figure*}

\vspace{2mm}
\noindent\textbf{Selection of comparison methods.} 
We choose four methods with accessible code for comparison. 
SpikeMS~\citep{eventmos:parameshwara_spikems_IROS_2021} is a learning-based model utilizing an encoder-decoder SNN architecture. %
Due to the absence of training code, we evaluated SpikeMS directly on the EVIMO dataset using their pre-trained weights provided by the authors. 
Since it exclusively segments AMOs, we calculate only its foreground segmentation result. 
EMSGC~\citep{eventmos:zhou_EMSGC_TNNLS_2021} is a traditional optimization method that provides hyperparameters for two subsequences only. %
We thus conduct subsequent comparisons with EMSGC and SpikeMS solely on these two sequences for fairness. 
IDOL~\citep{vis:wu_defenseIDOL_ECCV_2022} is a video instance segmentation method for video frames. %
We fine-tune IDOL on the EVIMO and EKubric datasets,  respectively, disregarding object semantic categories and retaining only the ``moving'' category to align the MOS objective. 
ZBS~\citep{cd:an_zbs_CVPR_2023} is a background subtraction method pretrained on large-scale data which only outputs the foreground mask, and we integrate its intermediate results to implement InsMOS. %
Additionally, we disregard category information from its built-in detector as static objects are excluded in its Background Modeling process.

It is worthwhile to emphasize that InsMOS is an emerging task and lacks standardized benchmarks. 
Some related MOS methods, including event-based motion segmentation~\citep{eventmos:chen_progressivemotionseg_AAAI_2022, eventmos:huang_progressive_CVPR_2023,
eventmos:Stamatios_GeneralMOS_3DV_2024}
and optical flow-based motion segmentation~\citep{mos:Meunier_EMMOS_TPAMI_2023}, are not compared here due to the unavailability of publicly accessible code or pretrained models. 
We retrain the VOS method IDOL and adopt the large-scale dataset pretrained method ZBS to achieve InsMOS and thus support MOS comparisons. 
In addition, we perform expanded comparisons on the BGS task to further demonstrate the performance of our method. 
We elaborate and analyze the comparison details and results below. 

\subsection{Evaluation Results}

\noindent\textbf{Comparison on the real world dataset.}
We first evaluate our model on the commonly used real-world MOS dataset EVIMO~\citep{eventdatasets:mitrokhin_EVIMO_IROS_2019}.
We select two recent event-only MOS methods with publicly available code for comparison, as no subsequent works provide publicly available code or model. 
Since EMSGC~\citep{eventmos:zhou_EMSGC_TNNLS_2021} only provides hyperparameters for the \textit{box\_00} and \textit{table\_01} sequences, we conduct a fair comparison with these two methods on these two sub-sequences at the top of Tab.~\ref{tab:evimo}. 
The observed performance gain ($\Delta$13.2~$\rm {mAP}$) illustrates the importance of introducing conventional image with events for MOS, highlighting the superiority of this combined setting over using events alone. 
Additionally, we compare our model with two recent frame-based methods, IDOL~\citep{vis:wu_defenseIDOL_ECCV_2022} and ZBS~\citep{cd:an_zbs_CVPR_2023}. 
At the bottom of Tab.~\ref{tab:evimo}, we improve over $\Delta$3.6~$\rm {mAP}$ on the full EVIMO eval set,  indicating that our model can extract effective motion information from events compared to using images alone. 
Furthermore, the lightweight version of our model (Ours-small) achieves superior results with shorter runtime.

\begin{table*}[tbp]
\caption{\textbf{Performance comparisons on the test set of the high-resolution complex EKubric~\citep{eventflow:Wan_RPEFlow_ICCV_2023} dataset}. 
We do not compare event-only methods because no training code is available. 
}
\centering
\setlength{\tabcolsep}{8pt}{
\begin{tabular}{ccccccc}
    \toprule
    Input & Method & $\rm {mAP}$ \Fix{$\uparrow$} & $\rm {mIoU}_{ins}$ \Fix{$\uparrow$} & $\rm {mIoU}_{01}$ \Fix{$\uparrow$} & Time & Param \\ %
    \midrule
    \multirow{2}*{$\!I_{t},\!I_{t\!+\!\Delta}\!$}
    & IDOL & 35.9 & 69.84 & 76.41 & 107.ms & 21.34M \\ 
    & ZBS & 34.3 & 68.27 & 74.59 & 110.ms & 175.4M \\ 
    \midrule
    \multirow{2}*{$\!I_t, \!E_{t:t\!+\!\Delta}$}
    & Ours & \textbf{44.6} & \textbf{79.66} & \textbf{82.65} & 76.2ms & 41.99M \\
    & Ours-small & 38.1 & 71.82 & 80.06 & \textbf{30.7}ms & \textbf{8.73}M \\ 
    \bottomrule
\end{tabular}
}
\label{tab:ekubric}
\end{table*}

\begin{figure*}[t]
\small
\centering
\includegraphics[width=\textwidth]{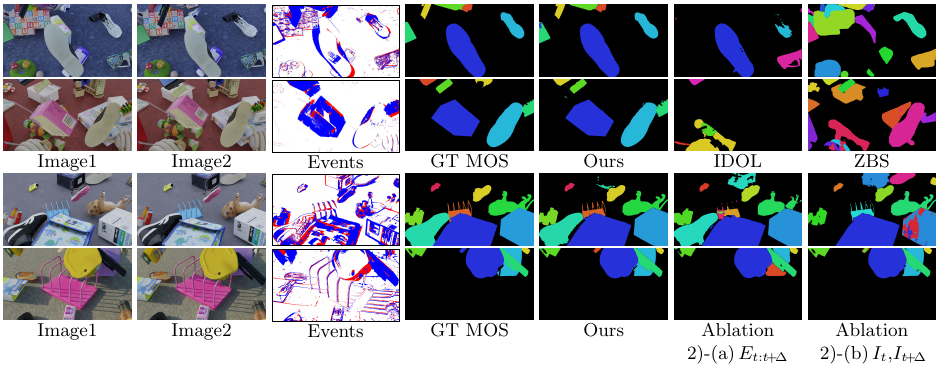}
\caption{\textbf{Visual comparisons on the simulated EKubric~\citep{eventflow:Wan_RPEFlow_ICCV_2023} dataset}. 
We compare two image-based methods, IDOL and ZBS, in the upper two samples, and two ablation models, replacing the input data with only events or images, in the bottom two samples.
} 
\label{viz:ekubric}
\end{figure*}

In addition, we also perform visual comparisons in Fig.~\ref{viz:evimo}. 
EMSGC~\citep{eventmos:zhou_EMSGC_TNNLS_2021} %
as an event-only method %
has difficulty separating neighboring objects due to limited texture information.
As a two-frame method, it is difficult for IDOL~\citep{vis:wu_defenseIDOL_ECCV_2022} to accurately determine whether an object is moving when accompanied by camera motion.
ZBS~\citep{cd:an_zbs_CVPR_2023} establishes associations for multi-frame segmentation based on object detection, thus it has a high false positive rate due to it might over-segment the scene. 
These observations align with previously highlighted limitations of relying solely on images or events.
Our framework can take advantage of the image and events to accurately segment each moving object, especially when they are close together. 
Our framework leverages both images and events to effectively segment each moving object, particularly in scenarios where objects are closely moving. 
In addition, the predicted optical flow also corresponds well with the ground-truth moving objects, demonstrating that incorporating optical flow guidance enables the learning of reasonable object motion in an unsupervised manner.
\begin{table*}[htbp]
\caption{\Fix{\textbf{Ablation experiments on model structures and training setups}. }
All of these models are trained and evaluated on the training and test sets of the EKubric dataset, respectively. 
The underlined items represent our final framework. 
}
\centering
\setlength{\tabcolsep}{6pt}{
\begin{tabular}{lcc|cccc}
    \toprule
    \multicolumn{1}{c}{\multirow{2}{*}{Ablation}} & \multicolumn{2}{c|}{\multirow{2}{*}{Model}} & \multicolumn{2}{c}{\multirow{1}{*}{InsMOS}} & \multicolumn{2}{c}{\multirow{1}{*}{Optical Flow}} \\
    \cmidrule{4-7}
    & & & $\rm {mAP}$ \Fix{$\uparrow$} & $\rm {mIoU}_{ins}$ \Fix{$\uparrow$} & $\rm {EPE}$ \Fix{$\downarrow$} & $\rm {1px}$ \Fix{$\uparrow$} \\ 
    \midrule
    \multirow{7}*{1)~model structure}
    & \underline{(a)} & \underline{\Fix{complete}~(CMA+FFE+CFL)} & \underline{44.6} & \underline{79.66} & \underline{2.39} & \underline{56.16\%} \\
    & (b) & CMA+FFE & 42.8 & 79.09 & 2.68 & 53.85\% \\
    & (c) & CMA+CFL & 38.5 & 77.80 & - & - \\
    & (d) & CMA & 36.4 & 77.02 & - & - \\
    & \Fix{(e)} & \Fix{FFE} & \Fix{35.3} & \Fix{76.41} & \Fix{2.89} & \Fix{50.12\%} \\
    & \Fix{(f)} & \Fix{CFL} & \Fix{33.8} & \Fix{75.06} & - & - \\
    & \Fix{(g)} & basic & 32.4 & 74.12 & - & - \\
    \midrule
    \multirow{3}*{2)~input data}
    & (a) & $\!E_{t:t\!+\!\Delta}$& 32.9 & 70.28 & 3.14 & 47.80\% \\
    \multirow{3}*{\Fix{based on complete model}}
    & (b) & $\!I_{t}, I_{t\!+\!\Delta}\!$ & 35.4 & 74.67 & 5.03 & 45.21\% \\
    & (c) & $\!I_{t}, F_{t:t\!+\!\Delta}^{\text{RAFT}}\!$ & 34.8 & 73.01 & 4.27 & 41.45\% \\
    & \underline{(d)} & \underline{$\!I_t, \!E_{t:t\!+\!\Delta}\!$} & \underline{44.6} & \underline{79.66} & \underline{2.39} & \underline{56.16\%} \\
    \midrule
    \multirow{2}*{\Fix{3)~input data}}
    & \Fix{(a)} & \Fix{$\!E_{t:t\!+\!\Delta}$} & \Fix{23.9} & \Fix{65.90} & - & - \\
    \multirow{2}*{\Fix{based on basic model}}
    & \Fix{(b)} & \Fix{$\!I_{t}, I_{t\!+\!\Delta}\!$} & \Fix{30.1} & \Fix{69.29} & - & - \\
    & \underline{\Fix{(c)}} & \underline{\Fix{$\!I_t, \!E_{t:t\!+\!\Delta}\!$}} & \underline{\Fix{32.4}} & \underline{\Fix{74.12}} & - & - \\
    \midrule
    \multirow{2}*{4)~backbone}
    & \underline{(a)} & \underline{MobileNet} & 38.1 & 71.82 & 3.05 & 47.41\% \\
    \multirow{2}*{\Fix{based on complete model}}
    & \underline{(b)} & \underline{ResNet50} & \underline{44.6} & \underline{79.66} & \underline{2.39} & \underline{56.16\%} \\
    & (c) & ResNet50+DCN & 45.9 & 80.20 & 2.21 & 58.64\% \\
    \midrule
    \multirow{2}*{\Fix{5)~augmentation direction in CMA}}
    & \Fix{(a)} & \Fix{texture} & \Fix{40.9} & \Fix{78.36} & \Fix{2.97} & \Fix{50.32\%} \\
    \multirow{2}*{\Fix{based on complete model}}
    & \Fix{(b)} & \Fix{motion} & \Fix{43.2} & \Fix{79.20} & \Fix{2.48} & \Fix{55.63\%} \\
    & \Fix{\underline{(c)}} & \Fix{\underline{both}} & \Fix{\underline{44.6}} & \Fix{\underline{79.66}} & \Fix{\underline{2.39}} & \Fix{\underline{56.16\%}} \\
    \midrule
    \multirow{2}*{6)~feature loss $L_{cl}$ \Fix{in CFL} }
    & \Fix{(a)} & \Fix{w/o whole $L_{cl}$} & 42.8 & 79.09 & 2.68 & 53.85\% \\
    \multirow{2}*{\Fix{based on complete model}}
    & (b) & w/o cross-sim \Fix{in $L_{cl}$} & 43.7 & 79.38 & 2.57 & 54.16\% \\
    & \underline{(c)} & \underline{w/ cross-sim \Fix{in $L_{cl}$}} & \underline{44.6} & \underline{79.66} & \underline{2.39} & \underline{56.16\%} \\
    \midrule
    \multirow{3}*{7)~ flow loss $L_{f}$ \Fix{in FFE} }  
    & (a) &  w/o \Fix{whole $L_{f}$} & 38.5 & 77.80 & - & - \\
    \multirow{3}*{\Fix{based on complete model}}
    & (b) & $L_{uf}$ & 43.0 & 78.25 & 5.49 & 50.66\% \\
    & \underline{(c)} & \underline{$L_{sf}$} & \underline{44.6} & \underline{79.66} & \underline{2.39} & \underline{56.16\%} \\
    & (d)& $L_{sf} \! + \! L_{uf}$ & 45.8 & 80.62 & 2.45 & 54.93\% \\
    \midrule
    \multirow{3}*{\Fix{8)~training frames $K$ in CFL}}
    & \Fix{(a)} & \Fix{$K\!=\!1$ (w/o CFL)} & \Fix{42.8} & \Fix{79.09} & \Fix{2.68} & \Fix{53.85\%} \\
    \multirow{3}*{\Fix{based on complete model}}
    & \Fix{\underline{(b)}} & \Fix{\underline{$K\!=\!2$}} & \Fix{\underline{44.6}} & \Fix{\underline{79.66}} & \Fix{\underline{2.39}} & \Fix{\underline{56.16\%}} \\
    & \Fix{(c)} & \Fix{$K\!=\!3$} & \Fix{45.0} & \Fix{79.81} & \Fix{2.31} & \Fix{56.65\%} \\
    & \Fix{(d)} & \Fix{$K\!=\!4$} & \Fix{45.0} & \Fix{79.86} & \Fix{2.28} & \Fix{56.83\%} \\
    \midrule
    \multirow{2}*{\Fix{9)~training batch size $bs$ and GPUs}}
    & \Fix{\underline{(a)}} & \Fix{\underline{$bs\!=\!16$, GPUs$=\!8$}} & \Fix{\underline{44.6}} & \Fix{\underline{79.66}} & \Fix{\underline{2.39}} & \Fix{\underline{56.16\%}} \\
    \multirow{2}*{\Fix{based on complete model}} 
    & \Fix{(b)} & \Fix{$bs\!=\!8$, GPUs$=\!4$ (2X iters)} & \Fix{44.4} & \Fix{79.62} & \Fix{2.42} & \Fix{56.01\%} \\
    & \Fix{(c)} & \Fix{$bs\!=\!4$, GPUs$=\!2$ (4X iters)} & \Fix{44.1} & \Fix{79.53} & \Fix{2.48} & \Fix{55.72\%} \\
    \bottomrule
\end{tabular}
}
\label{tab:ablation1}
\end{table*}

\vspace{2mm}
\noindent\textbf{Comparison on the high-resolution synthetic dataset.}
Due to the difficulty of capturing the ground-truth motion of multiple IMOs in real scenes, we further verify the performance of our model on the simulated high-resolution EKubric~\citep{eventflow:Wan_RPEFlow_ICCV_2023} dataset with more moving objects and more complex movements. 
Since there were no available training codes for SpikeMS or hyperparameters for EMSGC, we only compared them with two frame-based MOS methods. 
The quantitative comparisons in Tab.~\ref{tab:ekubric} clearly show that our method can locate and segment IMOs more accurately compared to methods that do not use events.
In addition, we also conduct visual comparisons in Fig.~\ref{viz:ekubric}.
By comparing with two image-based methods (the top two comparisons), we found that it is difficult to accurately determine whether an object is moving or not using only images, and further supported by the results of the image-only ablation model 2-(b). 
It is difficult to segment fine masks due to the lack of dense semantic information for the event-only ablation model 2-(a).
Nonetheless, the judgment of moving objects is more accurate, especially when the camera is moving at the same time.
Our framework effectively leverages the strengths of both modalities, yielding more accurate instance-level segmentation maps. 

\subsection{Ablation Studies}

To verify the effectiveness of each model component as well as the input settings in our proposed framework, we perform ablation experiments and analyze them individually.

\vspace{2mm}
\noindent\textbf{Network components.}
\Fix{To solve the instance-level MOS task in the simplest framework, both the feature enhancement structure and the contrastive feature learning objective are not necessary as their purpose is only to enhance feature learning.
Furthermore, we also verified in the ablation that the goal of segmenting moving objects can be achieved using only two frames or using only events as these settings can provide usable, although imperfect, spatio-temporal motion information. 
However, we need to specify motion confidence for each segmentation to adapt to varying numbers of objects, therefore it is necessary to build a new instance-level MOS framework with mask segmentation and motion classification decoders. 
Our second contribution of the instance-level segmentation framework is therefore necessary to build a minimal model (basic model).
We keep the mask and motion embedding segmentation steps outlined in Sec.~\ref{sec:decoder} as the 1)-(g) basic model architecture in Tab.~\ref{tab:ablation1}, which simply concatenates the event feature with the image feature. 
We can then build on this minimal model by adding components one by one to verify the effectiveness of each model structure or learning strategies proposed in our framework. 
}
The ablation comprises three components: cross-modal masked attention augmentation (CMA), contrastive feature learning (CFL), and flow-guided motion feature enhancement (FFE), integrated into models 1)-(a) to \Fix{1)-(f)} in Tab.~\ref{tab:ablation1}. 
\Fix{In short, the model that combines all the components achieves the best performance, which is our final complete model. }

\begin{figure*}[tbp]
\small
\centering
\includegraphics[width=0.8\textwidth]{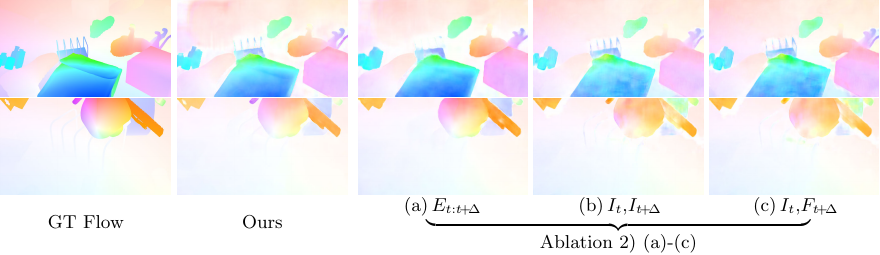}
\caption{
\textbf{Optical flow visualization} corresponding to the two samples in the bottom of Fig.~\ref{viz:ekubric}.
} 
\label{viz:flow_viz}
\end{figure*}

Specifically, comparing model 1)-(g) to 1)-(d), the inclusion of interaction between events and images significantly enhances performance, suggesting that information embedded in events and conventional images mutually benefits MOS. 
Comparison between 1)-(g) to 1)-(f) and 1)-(d) to 1)-(c) demonstrates the introduction of contrastive learning constraints on augmented features, facilitating the acquisition of more representative features and positively impacting segmentation. 
Conversely, the transition from 1)-(g) to 1)-(e) and 1)-(d) to 1)-(b) reveals a more substantial performance improvement upon integrating flow-guided motion feature learning.
\Fix{
We argue that the objective of contrastive feature learning is to ensure complementary representations of the two features, which is worthwhile for improving MOS performance. 
However, this process does not inherently capture specific roles, \ie, cannot accurately separate the required texture information and motion information for decoders, hence the improvement when just CFL alone is not significant.
In contrast, optical flow is treated as a concise and direct motion representation that can be used to guide feature learning toward effectively utilizing meaningful dense motion information from inputs. 
While motion information is explicitly guided, contrastive feature learning can then better model complementary correlations, leading to valuable context information. 
This is why combining CFL and FFE still improves the MOS performance even further even though the introduction of FFE is significant.
}
The combination of all three modules in model 1)-(e) achieves the best performance, elucidating the contribution of all proposed modules. 

\noindent\textbf{\Fix{Input modalities.}}
The comparison with existing methods above has shown that both event and image data are helpful for MOS. 
In order to further conduct ablation analysis, we replace the input data with only events and two images, which means that semantic and motion information needs to be learned from a single modality. 
In addition, we conducted visual comparisons of the MOS results as shown in Fig.~\ref{viz:teaser} and Fig.~\ref{viz:ekubric}. 
We find that using only events (\ie, Tab.~\ref{tab:ablation1} 2-(a)) yields a high accuracy of optical flow estimation, but it is still slightly inferior compared to using two frames (\ie, Tab.~\ref{tab:ablation1} 2-(b)) for our segmentation goal. 
Motion cannot be reliably modeled when only images are used, and thus moving objects cannot be accurately segmented.
Although better modeling of the motion, it is still difficult to achieve accurate and dense MOS since using only events lacks texture information.
We present the results of the optical flow estimation for these three models in Fig.~\ref{viz:flow_viz}, which also verifies this observation. 
Our proposed input setup combines the strengths of both images and events, maintaining a significant advantage in a fair comparison using the same segmentation framework.

We also try another setup that combines images with optical flow, which is utilized in many VOS methods~\citep{vos:zhao_adaptivezvos_ijcv_2024}. 
We pre-compute the optical flow using RAFT~\citep{flow:Teed_RAFT_ECCV_2020}, and replace the event inputs with computed optical flow while maintaining the original structure and learning strategy. 
Comparing models 2-(c) and 2-(d) in Tab.~\ref{tab:ablation1}, we observed that using events as the motion information source achieves better performance. 
This outcome can be attributed to the inherent uncertainty in the pre-computed optical flow, and the model makes it easier to learn motion information from events. 
We believe combining events and images captured by sensors is more practical. 

\Fix{Furthermore, we conduct the above experiments with different input modalities on the basic model without several feature fusion and learning components proposed in this paper.
Specifically, we evaluate the segmentation performance of different input modalities based on this basic model, which is devoid of optical flow estimation capability. 
In Tab.~\ref{tab:ablation1} 3)-(a) to (c), the experimental observations based on the basic model are consistent with those based on the complete model, \ie, combining image and events gives optimal benefit to segmentation performance even without our proposed feature fusion and learning strategies.
}

\vspace{2mm}
\noindent\textbf{Backbone.}
To illustrate the adaptability of our proposed framework to the backbone, we also experiment with three different backbone structures, MobileNet~V2~\citep{sandler2018mobilenetv2}, ResNet50~\citep{he2016deep}, and ResNet50+DCN~\citep{zhu2019deformable}.
The former channel of embeddings is $c \!=\! 128$ and the latter two are $c \!=\! 256$. 
In Tab.~\ref{tab:ablation1} 4)-(a) to (c), we find that better performance can be achieved with a more complex backbone. 
Since mainstream methods commonly use MobileNet and ResNet50, we denote the former model as ``Ours-small'' and the latter model as ``Ours''.

\begin{table*}[htbp]
\caption{\Fix{\textbf{Ablation experiments on inference setups}. 
All experiments are performed based on the pretrained complete model by adjusting the inference hyperparameters.
The underlined items represent our final choice of hyperparameters. 
}}
\centering
\setlength{\tabcolsep}{8pt}{
\begin{tabular}{lcc|cccc}
    \toprule
    \multicolumn{1}{c}{\multirow{2}{*}{Ablation}} & \multicolumn{2}{c|}{\multirow{2}{*}{Model}} & \multicolumn{2}{c}{\multirow{1}{*}{InsMOS}} & \multicolumn{2}{c}{\multirow{1}{*}{Optical Flow}} \\
    \cmidrule{4-7}
    & & & $\rm {mAP}$ \Fix{$\uparrow$} & $\rm {mIoU}_{ins}$ \Fix{$\uparrow$} & $\rm {EPE}$ \Fix{$\downarrow$} & $\rm {1px}$ \Fix{$\uparrow$} \\ 
    \midrule
    \multirow{4}*{7)~moving threshold $\theta$} & (a) & {\small $\theta\!=\!0.1$} & 38.7 & 79.39 & 2.39 & 56.16\% \\
    & \underline{(b)} & \underline{{\small $\theta\!=\!0.3$}} & 44.6 & \underline{79.66} & 2.39 & 56.16\% \\
    & (c) & {\small $\theta\!=\!0.5$} & \underline{46.3} & 75.40  & 2.39 & 56.16\% \\
    & (d) & {\small $\theta\!=\!0.7$} & 44.4 & 71.70 & 2.39 & 56.16\% \\
    \midrule
    \multirow{3}*{8)~inter-frame inference}
    & (a) & {\small $\!I_t \!+\! \!E_{t:t\!+\! {\Delta \!/\! 3}}\!$} & 40.6 & 76.17 & - & - \\
    & (b) & {\small $\!I_t \!+\! \!E_{t:t\!+\! {\Delta \!/\! 2}}\!$} & 44.2 & 78.71 & - & - \\
    & \underline{(c)} & \underline{\small $\!I_t \!+\! \!E_{t:t\!+\! {\Delta}}\!$} & \underline{44.6} & \underline{79.66} & \underline{2.39} & \underline{56.16\%} \\
    \bottomrule
\end{tabular}
}
\label{tab:ablation2}
\end{table*}

\vspace{2mm}
\noindent\textbf{Augmentation direction.} 
In Tab.~\ref{tab:ablation1} 2-(a), we verify that the InsMOS goal can be achieved by using only events, as the event data is sufficient to classify whether an object is moving or not. 
We go further and conduct three experiments focused on the feature augmentation directions in CMA, \ie, texture-only augmentation (inputting event feature as motion feature to the motion classifier), motion-only augmentation (inputting image feature as texture feature for mask segmentation), and dual branches augmentation. 
The results in Tab.~\ref{tab:ablation1} 5-(a) to (c) illustrate the superiority of augmenting both texture and motion features simultaneously from a single image and events. 
If we only augment the texture feature, the motion branch with only events cannot capture sufficient dense spatial information. 
If we only augment the motion feature, the semantic branch lacks information such as complementary motion contours from events, which affects the mask segmentation step and thus the output MOS accuracy. 
Simultaneous augmentation of texture and motion features can more adequately exploit information from the input image and event data beneficial for both mask segmentation, motion classification, and optical flow estimation. 

\vspace{2mm}
\noindent\textbf{Training loss.} 
We also ablate the configuration of the loss function for explicit feature learning. 
The contrastive feature learning loss $L_{cl}$ in Eq.~\eqref{eq:clloss} aims to reduce the similarity of texture and motion features, thus cross-similarity term $\! \sum_{k_1\!, k_2}^{K} \! CS_{T_{k_1}\!, M_{k_2}}$ plays a crucial role in minimizing objective.
In Tab.\ref{tab:ablation1} 6-(b), removing the cross-similarity term in the CMA process leads to performance degradation. 
The flow losses $L_{sf}$ and $L_uf$ in Eq.~\eqref{eq:flowsf} and Eq.~\eqref{eq:flowuf} aim to constrain the learning of motion information from event data. 
The choice between using the self-supervised former or the supervised latter depends on whether the dataset has reliable ground-truth optical flows. 
We conduct combinatorial experiments on the EKubric dataset, which has high-quality simulated optical flow annotations. 
The results in Tab.\ref{tab:ablation1} 7-(a), (b) and (c) show that both supervised and unsupervised training can improve the performance of MOS compared to not enabling the optical flow enhancement learning module. 
On this basis, combining $L_{sf} \!+\! L_{uf}$, 5-(d) %
would further benefit the MOS, but with a slight degradation of flow accuracy due to uncertainties of photometric consistency in unsupervised flow loss. 

\vspace{2mm}
\Fix{
\noindent\textbf{Training frames.} 
As described in the above training details, the number of frames in multi-frame contrastive feature learning during training is set to $K = 2$, and the batch size is set to $16$ (each GPU is $2$). 
The trade-off between batch size and the number of frames is limited by the 24G memory of the RTX 3090 GPU. 
We add an entry named ``8)~training frames $K$ in CFL'' to Tab.~\ref{tab:ablation1} to discuss this issue specifically. 
It is worth re-emphasizing that to ensure a fair comparison, we only supervise the segmentation outputs for the first frame as a data term, while the remaining multiple frames are only used for contrastive feature learning as a regularization term. 
The $K=1$ case is actually an ablation variant of removing the contrastive feature learning (CFL) strategy. 
From the experimental results in Tab.~\ref{tab:ablation1} 8-(a) to (d), taking more frames to build more pairs of positive and negative samples facilitates the learning of discriminative texture and motion information, which consequently improves both moving object segmentation and optical flow estimation performance, but also introduces more training computation consumption.
We found that more than two frames bring decreasing improvements, so we still keep $K=2$ as our final model. 
}

\vspace{2mm}
\Fix{
\noindent\textbf{Training batch size and GPUs.} 
A well-recognized practice is using a large batch size in contrastive learning to achieve good downstream performance~\citep{learning:chen2020simple, learning:tsai2021selfsupervised}.
Since our work takes a new two-modal setup with more input data, as well as introduces the cross-batch and multi-frame based contrastive feature learning strategy that requires a larger batch size, both of these lead to more memory and GPU requirements than existing methods. 
The trade-off between batch size and the number of frames is limited by the 24G memory of the RTX 3090 GPU.
Specifically, we need 8$\times$24G to set the batch size to 16 on the EKubric dataset with a resolution of 1280$\times$720, while 4$\times$24G is needed for 346$\times$260 at EVIMO. 
To verify the impact, we experiment with reducing the batch size and the number of GPUs by half, but training the same number of epochs (twice training iterations), resulting in a slight performance loss on the EKubric dataset in Tab.~\ref{tab:ablation1} 9)-(a) to (c). 
We argue that smaller batch sizes undermine the ability of contrastive feature learning to produce diverse positive and negative samples, which consequently impacts the segmentation performance. 
So we keep training on the Ekubric dataset with 8 GPUs. 
It is worth noting that although a large batch size is required for training, it does not affect the high efficiency of our model in inference on a single GPU, which we highlight in Sec.~\ref{sec:efficiency}. 
}

\begin{figure*}[tb]
\centering
\includegraphics[width=\textwidth]{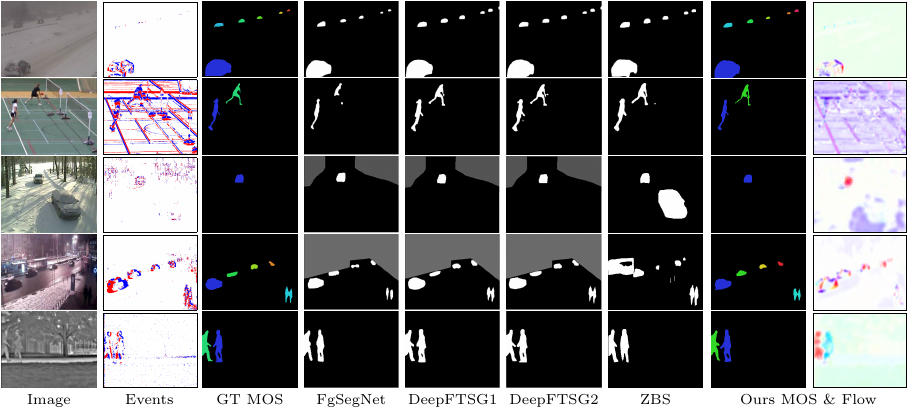}
\caption{\textbf{Visual comparisons on the CDNet2014~\citep{cd:wang_cdnet_CVPRW_2014} dataset}. 
From top to bottom, the examples are from sequences ``intermittentObjectMotion/winterDriveway'', ``cameraJitter/badminton'', ``badWeather/blizzard'', ``nightVideos/tramStation'', and ``thermal/park''. 
\Fix{To match visualizations in the DeepFTSG~\citep{cd:rahmon_deepftsg_IJCV_2024} paper, the methods we compare are all visualized by adopting the setting of only segmenting a foreground mask as defined by the BGS task, only the GT and Ours visualizations are color-coded that reflect the capability of instance-level segmentation. 
The gray areas in some results are the uncertainty areas given by the raw dataset, which we omit for brevity. 
}
} 
\label{viz:cdnet}
\end{figure*}

\vspace{2mm}
\noindent\textbf{Moving threshold.} 
The moving confidence score threshold $\theta$ is a predefined hyperparameter that directly impacts the selection of valid masks from segmentation embeddings, but does not affect the optical flow estimation branch. 
To assess its influence, we perform ablations using the same pre-trained model. 
Evaluation results in Table \ref{tab:ablation2}~10)-(a) to (d) indicate that different $\theta$ leads to notable effects on both segmentation metrics. 
Higher thresholds may exclude true masks, leading to a significant decrease in $\rm {mIoU}_{ins}$. 
However, since they also exclude false predictions, the $\rm {mAP}$ results may not decrease correspondingly but might potentially increase. 
Conversely, lower thresholds introduce more incorrect predictions, resulting in a significant decrease in $\rm {mAP}$. 
Thus we set $\theta\!=\!0.3$ to strike a balance between the two metrics. 

\begin{table*}[tb]
    \begin{center}
    \caption{
    \textbf{Overall and per-category F-Measure \Fix{$\uparrow$} comparisons on the CDNet2014~\citep{cd:wang_cdnet_CVPRW_2014} dataset.}
    Our model notably improves performance, particularly with large motion (dynamicBackground, intermittentObjectMotion, lowFramerate) and night scenes (nightVideos). 
    }
    \begin{tabular}{c|ccccccccccc|c}
        \toprule
        Method  & baseline  & camjitt   & dynbg  & intmot & shadow  & thermal   & badwea  & lowfr   & night  & PTZ   & turbul  & Overall  \\
        \midrule
        FgSegNet &   {0.6926} &  {0.4266}&   {0.3634}&   {0.2002}&   {0.5295}&   {0.6038}&   {0.3277}&   {0.2482}&   {0.2800}&   {0.3503}&   {0.0643}&   {0.3715} \\
        ADNN-IB & 0.9522 & 0.8245 & 0.9166 & 0.6978 & 0.9054 & 0.8058 & 0.8245 & 0.7028 & 0.4872 & 0.2931 & 0.7620 & 0.7451 \\
        BSUV-Net & \textbf{0.9693} & 0.7743 & 0.7967 & 0.7499 & 0.9233 & 0.8581 & 0.8713 & 0.6797 & 0.6987 & 0.6282 & 0.7051 & 0.7868 \\
        BSUV-Net 2.0 & 0.9620 & 0.9004 & 0.9057 & 0.8263 & 0.9562 & 0.8932 & 0.8844 & 0.7902 & 0.5857 & 0.7037 & 0.8174 & 0.8387 \\
        ZBS & {0.9653} &   \textbf{0.9545}&   {0.9290}&   {0.8758}&   \textbf{0.9765}&   {0.8698}&   {0.9229}&   {0.7433}&   {0.6800}&   {0.8133}&   {0.6358}&   {0.8515} \\
        Ours & 0.9552 & 0.9342 & \textbf{0.9586} & \textbf{0.9203} & 0.9253 & \textbf{0.9610} & \textbf{0.9656} & \textbf{0.8053} & \textbf{0.8838} & \textbf{0.9023} & \textbf{0.8861} & \textbf{0.9180} \\
        \bottomrule
    \end{tabular}
    \label{tab:cdnet}
    \end{center}
\end{table*}

\subsection{Inter-Frame Inference}
Since our framework takes a single image with its following events as inputs, we can naturally adjust the period of input events flexibly.
In the above comparisons, we follow the common practice of inputting the events $E_{t:t\!+\! {\Delta}}$ between two consecutive frames, \ie, $\Delta$ equal to the shutter speed of the conventional frame-based camera. 
To exemplify the flexibility, we utilize a pretrained model with a fixed event duration of $\Delta$ and evaluate it with fewer events. 
In models 11)-(a) and (b) of Tab.~\ref{tab:ablation2}, we report the MOS performance with 1/3 and 1/2 inter-frame events, while not reporting the flow performance due to the absence of corresponding inter-frame optical flow ground-truth. 
The results indicate that even though the model has not been trained with inter-frame events, it still exhibits good motion identification capability. 
The slight performance degradation may stem from the difficulty in distinguishing small movements and non-linear motion being less evident in the earlier events. 
\Fix{This observation suggests that we do not need to capture the entire events between two frames, but only a portion is needed as the model input.
We believe this is sufficient to illustrate our flexibility for inter-frame inference and demonstrate the potential of our two-modal framework in escaping the frame rate limitation of images, which is not available in the existing setting of taking images only or combining images with optical flow. }

\subsection{Adapt to Background Subtraction Task}

We extend our analysis on the CDNet2014~\citep{cd:wang_cdnet_CVPRW_2014} dataset to provide intuitive comparisons on an image-based Background Subtraction (BGS) benchmark. 
The CDNet2014 dataset is a large-scale database for BGS and change detection tasks, which consists of 53 videos and 11 different categories with diverse image resolution ranging from 320$\times$420 to 480$\times$720. 
\Fix{To obtain instance-level segmentation annotations, we isolate foreground change regions using nearest-neighbor search and clustering strategies. } 
We then simulate events using vid2e~\citep{eventdatasets:Gehrig_VideoToEvent_CVPR_2020}. %
Following to FgSegNet~\citep{cd:lim_fgsegnet_PRL_2018}\footnote{\url{https://github.com/lim-anggun/FgSegNet/tree/master/FgSegNet\_dataset2014}}, we utilize up to 50 frames from each video as training data and the remaining frames as testing data. 
This yields $\sim$1,200 pairs of training data and $\sim$50,000 pairs of testing data. 
We take the same training settings as on the EVIMO dataset.

\begin{table*}[ht]
\caption{\Fix{\textbf{Efficiency analysis of each component, including Time~(ms) and Parameters~(M) with 1280$\times$720 inputs}. 
The time in ``Total'' is required for model inference, while the total number of parameters includes the layers in CFL and FFE that are disabled during inference.
We analyze the time and the number of parameters of our three variants, where the small model satisfies the real-time requirements in 1280$\times$720.}
}
\centering
\setlength{\tabcolsep}{1pt}{
\begin{tabular}{ccccccccccc}
\toprule
\multirow{2}*{Model} & \multicolumn{2}{c}{Backbone} & \multicolumn{2}{c}{CMA} & \multicolumn{2}{c}{Decode} & \multicolumn{2}{c}{Output} & \multicolumn{2}{c}{Total} \\ 
\cmidrule(r){2-3} \cmidrule(r){4-5} \cmidrule(r){6-7} \cmidrule(r){8-9} \cmidrule(r){10-11}
& Time~(ms) & Param~(M) & Time~(ms) & Param~(M) & Time~(ms) & Param~(M) & Time~(ms) & Param~(M) & Time~(ms) & Param~(M) \\ 
\midrule
\Fix{Images only} & \Fix{23.6} & \Fix{8.6} & \Fix{28.9} & \Fix{11.4} & \Fix{14.8} & \Fix{12.2} & \Fix{8.6} & - & \Fix{75.9} & \Fix{33.39} \\ 
\Fix{Events only} & \Fix{12.1} & \Fix{8.6} & \Fix{28.9} & \Fix{11.4} & \Fix{14.8} & \Fix{12.2} & \Fix{8.6} & - & \Fix{64.3} & \Fix{33.39} \\ 
\Fix{Basic model} & \Fix{23.9} & \Fix{17.2} & \Fix{-} & \Fix{-} & \Fix{14.8} & \Fix{12.2} & \Fix{8.6} & - & \Fix{47.3} & \Fix{29.46} \\ 
{Ours} & 23.9 & {17.2} & 28.9 & {11.4} & 14.8 & {12.2} & 8.6 & {-} & 76.2 & 41.99 \\ 
{Ours-small} & 9.6 & {3.6} & 9.1 & {1.3} & 6.4 & {2.7} & 5.6 & {-} & 30.7 & 8.73 \\ 
{Ours-DCN} & 29.4 & {18.2} & 28.9 & {11.4} & 14.8 & {12.2} & 8.6 & {-} & 81.7 & 42.98 \\ 
\bottomrule
\end{tabular}
}
\label{tab:runtime}
\end{table*}

In Tab.~\ref{tab:cdnet}, we compare five SOTA image-based BGS methods, FgSegNet~\citep{cd:lim_fgsegnet_PRL_2018}, ADNN-IB~\citep{cd:zhao_universalBGS_TIP_2022}, BSUV-Net~\citep{cd:tezcan_bsuv1_WACV_2020}, BSUV-Net 2.0~\citep{cd:tezcan_bsuv2_access_2021}, and ZBS~\citep{cd:an_zbs_CVPR_2023}, which are all using the same single model to evaluate on all CDNet sequences. 
Note that other methods, such as FgSegNet2~\citep{cd:lim_fgsegnetv2_PAA_2020} and DeepFTSG~\citep{cd:rahmon_deepftsg_IJCV_2024}, adopt the strategy of training different models on different sequences separately to obtain better results, and we do not compare them \Fix{quantitatively}. 
\Fix{We find that our method performs well on most sequences, but presents a noticeable disadvantage on the shadow sequence.
The compared methods use two or more images to exploit temporal information about the shadowed areas, which can mitigate the interference of shadows in segmenting foreground objects. 
However, we use a single image with subsequent events, and events perceive objects and shadows indiscriminately, which impacts the localization of moving objects in the absence of temporal information from images. 
We argue that this is why our method is worse than recent image-based methods in such challenging scenes. 
However, we believe that our results on most sequences can adequately illustrate the substantial benefits of combining image with events for MOS. }
We also provide visual comparisons in Fig.~\ref{viz:cdnet}, and our model achieves robust segmentation performance while enabling both instance-level segmentation and optical flow estimation of foreground objects over these BGS methods. 
The experimental results underscore the advantages that event data offers to the BGS task, particularly in challenging scenarios such as night and bad weather. 

\subsection{Efficiency Analysis}
\label{sec:efficiency}

We analyze the time consumption of each framework component in Tab.~\ref{tab:runtime}. 
The most time-consuming components are CMA and backbone because the former involves attention operations, and the latter has a large number of parameters. 
The process of output fusion produces the instance-level segmentation results. 
While this process has no learnable parameters, its time consumption depends on the number of embeddings that remain a non-negligible factor for inference. 
Overall, our framework demonstrates high efficiency, with both versions of our model meeting real-time requirements at DAVIS 346\footnote{\url{https://inivation.com/solutions/cameras/}} resolution (346$\times$260, the highest resolution commercially available event camera capable of simultaneously outputting conventional images), and the smaller version also satisfying at Sony IMX636\footnote{\url{https://www.sony-semicon.com/en/products/is/industry/evs.html}} resolution (1280$\times$720). 
This aligns with the expectations of low latency and fast response in event-based neuromorphic vision. 
\Fix{
Furthermore, We find that if we introduce common inference acceleration strategies (mixed-precision and torch compilation) for our normal version model, the time consumption can be further accelerated from 76.2ms ($\sim$ 13 fps) to 34.0ms ($\sim$ 29 fps). 
We believe that the above inference time consumption results show the high efficiency of our proposed framework toward practical deployment requirements. 
}

\Fix{
In addition, we perform efficiency analyses on the two unimodal models in Tab.~\ref{tab:ablation1} 2)-(a) and (b) to analyze the computation cost introduced by two modalities.
The encoder structure is basically the same for the event and image branches, with only the event voxelization preprocessing and the different numbers of input channels. 
So the time for a single run of both encoders is similar.
Notably, cross-modal attention in the two models degrades to self-attention, but the time consumption is unchanged because the feature dimensions remain the same. 
The difference is the encoders are executed twice in total in the two-frame based model and in our fusion model, whereas the event-only model requires only a single run. 
As shown in Tab.~\ref{tab:runtime}, our model takes essentially the same inference time as these two unimodal models, except the events-only model takes half time in the Backbone step since there is no image branch. 
In addition, we also add a test of the basic model in Tab.~\ref{tab:ablation1} 1)-(g), which reveals the difference is only the CMA module, since CFL and FFE are disabled in inference. 
The above results illustrate that the difference in computational complexity brought by different input modalities is small, and the effect of improvements on the basic framework is acceptable. 
}

\begin{figure}[tbp]
\centering
\includegraphics[width=0.80\linewidth]{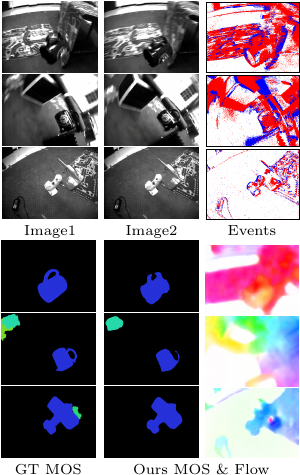}
\caption{
\textbf{Failure cases on the EVIMO~\citep{eventdatasets:mitrokhin_EVIMO_IROS_2019} dataset}. 
These cases include incomplete object outlines caused by poor image quality in reflective areas, objects situated at the edges of fast-moving scenes where determining their motion is difficult, and nearby objects with a significant size difference that might be missed.
} 
\label{viz:fail}
\end{figure}

\subsection{Failure Cases and Limitations}

We found some representative failure cases from the evaluation results on the real EVIMO dataset as visualized in Fig.~\ref{viz:fail}.
The listed cases include incomplete object outlines, misjudging edge objects, and missing neighboring objects. 
These are caused by data issues which indicates that such difficult cases are still challenging even if introducing events can significantly improve MOS performance.

\Fix{
Introducing new multi-modal settings will always require additional data collection requirements, but we believe this is worthwhile with the potential benefits of the data.
The commercial DAVIS camera can output event data with synchronized low-resolution images. 
However, no single sensor can simultaneously provide spatio-temporally aligned and high-resolution two modalities data. 
A common practice is building a dual-camera system (two separate different sensors with/without a beam splitter, sample model\footnote{\url{https://centuryarks.com/en/silkyevcam-bothview-c/}}) and performing post-alignment, which is frequently used in the academic community~\citep{eventapp:wang_jointfilter_CVPR_2020, eventapp:tulyakov_timelens_CVPR_2021, eventapp:tulyakov_timelenspp_CVPR_2022, eventapp:cho_noncoaxialdeblur_ICCV_2023, eventapp:kim_towardslowlight_ECCV_2024}. }
This may introduce alignment errors, which we plan to investigate in future work. 
\Fix{In addition, we currently provide thorough validation from theoretical analyses and experimental comparisons of the remarkable performance advantages for InsMOS introduced by fusing event and image data. 
We conclude that this new multi-modal setup is easy to deploy physically without costly data collection requirements, and brings significant benefits in spatio-temporally dense dynamic perceptions compared to uni-modal ones. 
Looking ahead, beyond our current focus on a minimal setup that fuses a single image with events (\ie, single images and their subsequent events), utilizing more images and events in a video streaming pipeline is a worthy future research topic.
}

\section{Conclusion}
\label{sec:conclusion}
In this paper, we propose the first InsMOS framework that integrates a single image and events to achieve instance-level and pixel-wise segmentation of all IMOs. 
Our proposed framework leverages both explicit and implicit fusion of event and image features to learn discriminative texture and motion features through cross-modal feature augmentation, contrastive feature learning, and flow-guided motion feature enhancement. 
Extensive experiments demonstrate that our framework is not only computationally efficient but also significantly outperforms existing MOS methods relying on a single modality. 
We emphasize the benefits of combining image and event data for MOS tasks, providing valuable insights into event-based motion-related vision. 

\section*{Declarations}

\noindent
\textbf{Acknowledgements.}
This research was supported in part by the National Natural Science Foundation of China (62271410, 62001394, 62401021), the Fundamental Research Funds for the Central Universities, the Agency for Science, Technology and Research (A*STAR) under its MTC Programmatic Funds (Grant No. M23L7b0021). 
Zhexiong Wan was also supported by the Program of China Scholarship Council (202306290193), and the Innovation Foundation for Doctor Dissertation of Northwestern Polytechnical University (CX2023013). 
Bin Fan was also supported by the China National Postdoctoral Program for Innovative Talents (BX20230013) and the China Postdoctoral Science Foundation (2024M750101).

\noindent
\textbf{Competing interests.}
The authors declare that they have no conflict of interest.

\noindent
\textbf{Data availability.}
All datasets used are publicly available. The source code of our model with model training details and pre-trained weights is released at \url{https://npucvr.github.io/EvInsMOS}.

\end{sloppypar}

\small{
    \bibliographystyle{spbasic}
    \bibliography{1_Event_Reference}
}

\end{document}